\documentclass[journal]{IEEEtran}

\usepackage{array}
\usepackage{booktabs} 
\usepackage{amsfonts}
\usepackage{amsmath,graphicx,setspace,algorithm,algorithmic,multirow,cases,amssymb,subfigure}
\usepackage{lettrine}
\usepackage{multirow}
\usepackage{xcolor}
\usepackage{times}
\usepackage{graphicx}
\usepackage{amsmath}
\usepackage{indentfirst}
\usepackage{booktabs}
\usepackage{blindtext}
\usepackage{array}
\usepackage{tabularx,booktabs}
\usepackage{color, soul}
\usepackage{bm}
\usepackage{cite}

\usepackage[
linkcolor=red,anchorcolor=blue,citecolor=green]{hyperref}
\usepackage{threeparttable}
\usepackage{float}
\newcommand{\RNum}[1]{\uppercase\expandafter{\romannumeral #1\relax}}

\begin{document}

\title{Bio-Inspired Representation Learning for Visual Attention Prediction}

\author{
        Yuan~Yuan,~\IEEEmembership{Senior Member,~IEEE,}
        Hailong~Ning,~\IEEEmembership{}and Xiaoqiang~Lu, \IEEEmembership{Senior Member,~IEEE}

\thanks{Manuscript received XX XX, XXXX.}

\thanks{
This work was supported in part by the National Natural Science Foundation of China under Grant 61806193, Grant 61702498, and Grant 61772510,
in part by the Young Top-notch Talent Program of Chinese Academy of Sciences under Grant QYZDB-SSW-JSC015, in part by the National Key R\&D Program of China under Grant 2017YFB0502900, in part by the CAS "Light of West China" Program under Grant XAB2017B26, and Grant XAB2017B15,
in part by the National Natural Science Found for Distinguished Young Scholars under Grant 61825603,
in part by the State Key Program of National Natural Science of China under Grant 61632018. \emph{(Corresponding author: Xiaoqiang Lu)}

Y. Yuan is with the Center for OPTical IMagery Analysis and Learning, School of the Computer Science, Northwestern Polytechnical University, Xi'an 710072, Shaanxi, P. R. China.

H. Ning is with the Key Laboratory of Spectral Imaging Technology CAS, Xi'an Institute of Optics and Precision Mechanics, Chinese Academy of Sciences, Xi'an 710119, Shaanxi, P. R. China, and also with the University of Chinese Academy of Sciences, Beijing 100049, P. R. China.

X. Lu is with the Key Laboratory of Spectral Imaging Technology CAS, Xi'an Institute of Optics and Precision Mechanics, Chinese Academy of Sciences, Xi'an 710119, Shaanxi, P. R. China (e-mail: luxq666666@gmail.com).}}
\maketitle

\begin{abstract}Visual Attention Prediction (VAP) is a significant and imperative issue in the field of computer vision. Most of existing VAP methods are based on deep learning. However, they do not fully take advantage of the low-level contrast features while generating the visual attention map. In this paper, a novel VAP method is proposed to generate visual attention map via bio-inspired representation learning. The bio-inspired representation learning combines both low-level contrast and high-level semantic features simultaneously, which are developed by the fact that human eye is sensitive to the patches with high contrast and objects with high semantics. The proposed method is composed of three main steps: 1) feature extraction, 2) bio-inspired representation learning and 3) visual attention map generation. Firstly, the high-level semantic feature is extracted from the refined VGG16, while the low-level contrast feature is extracted by the proposed contrast feature extraction block in a deep network. Secondly, during bio-inspired representation learning, both the extracted low-level contrast and high-level semantic features are combined by the designed densely connected block, which is proposed to concatenate various features scale by scale. Finally, the weighted-fusion layer is exploited to generate the ultimate visual attention map based on the obtained representations after bio-inspired representation learning. Extensive experiments are performed  to demonstrate the effectiveness of the proposed method.
\end{abstract}


\IEEEpeerreviewmaketitle

\section{Introduction}
\IEEEPARstart{W}{ith} the rapid development of mobile Internet, data are soaring, and the era of big data has come. We have to deal with a lot of information every day \cite{Zhangtcyb16, Yutcyb2017, Hantip13}, especially in the form of pictures and videos. Saliency detection provides us an effective idea for acquiring the most valuable information from massive data. The study of saliency methods can be divided into two research directions \cite{Han2016Two}: 1) the prediction of human eye fixation and 2) the salient object detection. The former is to predict the gaze positions that human focus on at first glance, attaching to a  regression problem. And the latter is to detect the salient objects in an observed scene, which is similar to hyperspectral image classification \cite{ZhangZDYT19}, belonging to a classification problem. In this paper, we focus on the eye fixation prediction task to get a continuous-valued density map, named visual attention map. The visual attention map has been used in scene understanding \cite{Zhang2017Revealing, Bo2014A, Yao2017Revisiting, Lu2017Semi}, object recognition \cite{Zhang2017Airport}, target detection \cite{Bo2014Target}, target tracking \cite{Zhang2017Online}, image quality assessment \cite{Jia2018Saliency} and video compression \cite{Hadizadeh2013Saliency}.

\begin{figure}
\begin{center}
\includegraphics[width=\linewidth]{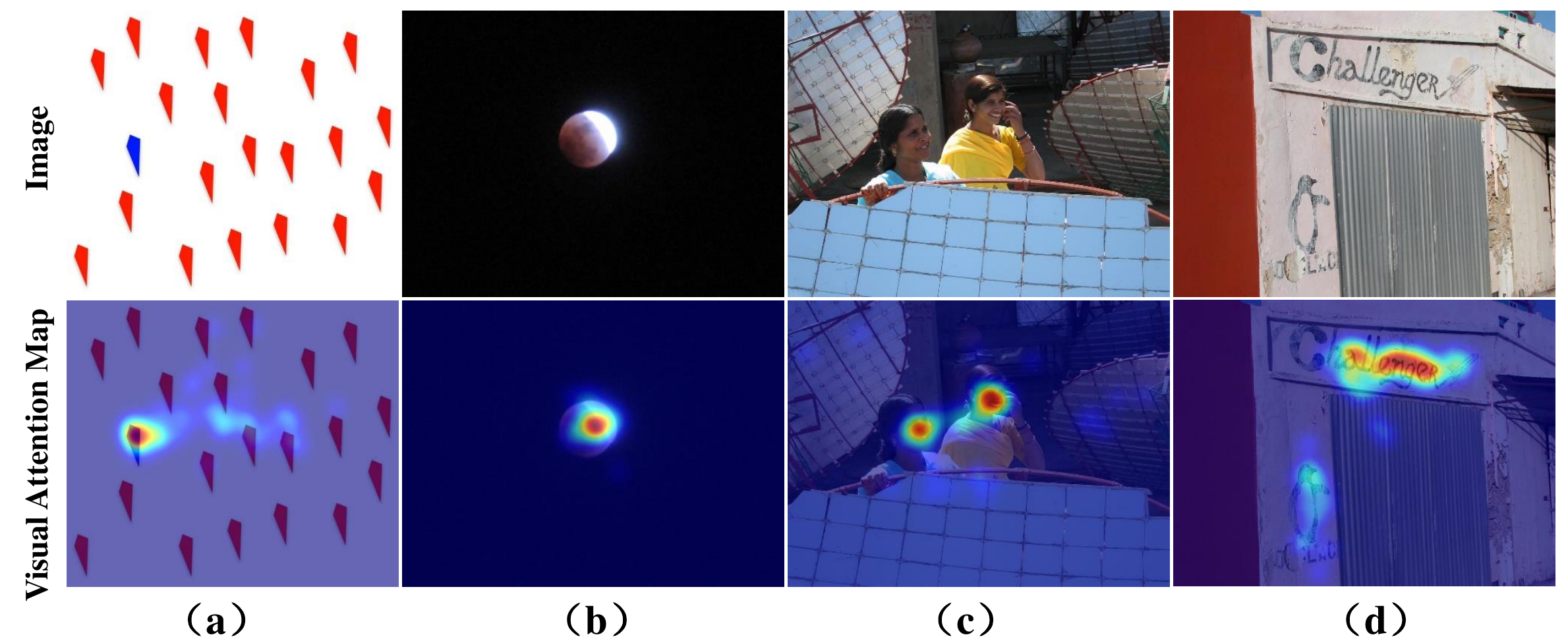}
\renewcommand{\figurename}{Fig.}
\vspace{-8 mm}
\end{center}
    \caption{\small{Various visual attention maps for different images in MIT1003 dataset: (a) and (b) are images containing high-contrast patches and corresponding visual attention maps, (c) and (d) are images containing high-semantics objects and corresponding visual attention maps.}}
\label{fig:1}
\end{figure}

In the field of the neuroscience, humans are more interested in the patch which stands out from its surrounding area. Fig. \ref{fig:1} shows four different images in MIT1003 dataset \cite{Judd2010Learning} and corresponding visual attention maps, which are generated by recording eye tracking data from 15 viewers when they observe these images freely. It is obvious to find that the human eye is sensitive to these patches with high-contrast, for instance, the blue diamond pattern in Fig. \ref{fig:1} (a) and the moon in Fig. \ref{fig:1} (b). At the same time, these objects with obvious semantic information is more easily to attract human eyes, such as the human faces in Fig. \ref{fig:1} (c) and the text in Fig. \ref{fig:1} (d). The reason for this fact is the unique physiological structure of human eyes. Relying on the unique physiological structure, humans can fixate their high-resolution fovea on things they want to see based on two attention mechanisms. The first attention mechanism is that human may only fixate the high-contrast patch at the first glance. The second attention mechanism is that human may notice some objects with obvious semantic features after rapid fixation. As a result, to introduce a superior visual attention computational method, we ought to take the two attention mechanisms into consideration simultaneously. In other words, both low-level contrast and high-level semantic features are supposed to be fully leveraged for VAP.

In the existing related works \cite{Itti2002A, Cheng2011Global, Itti2012Exploiting, Kummerer2017Understanding, Sss2017DeepFix}, the human attention in an image is described by two kinds of features: low-level contrast features and high-level semantic features. The low-level contrast features are developed to describe the discriminative regions in the image, because the contrast plays an important role in visual attention cognitive. In other words, high contrast regions are more easily noticed by people. On the contrary, the high-level semantic features are driven by internal semantic aspects (face, people, text, {\it etc.}). They can provide better locations about the semantic regions.
As a result, researchers in the field of computer vision have adopted these two kinds of features respectively for {\it Visual Attention Prediction} (VAP). The majority of earlier VAP methods \cite{Itti2002A, Cheng2011Global, Itti2012Exploiting} employ low-level contrast features. Specifically, color, intensity, orientation and other visual features are extracted manually to predict the visual attention map by calculating the contrast between the central area and the surrounding area. However, these low-level contrast features methods do not adequately take high-level semantic aspects in an observed scene into account. In recent years, the emerging deep {\it Convolutional Neural Networks} (CNNs) \cite{Zhong2017Computational} and the availability of large datasets have achieved impressive results in VAP. CNNs can extract powerful features to predict visual attention in an end-to-end manner. However, these CNNs methods focus on building the high-level semantic features from the last convolutional layers or fully-connected layers, making it hard to capture the low-level contrast features.

\begin{figure*}
\begin{center}
\includegraphics[width=\linewidth]{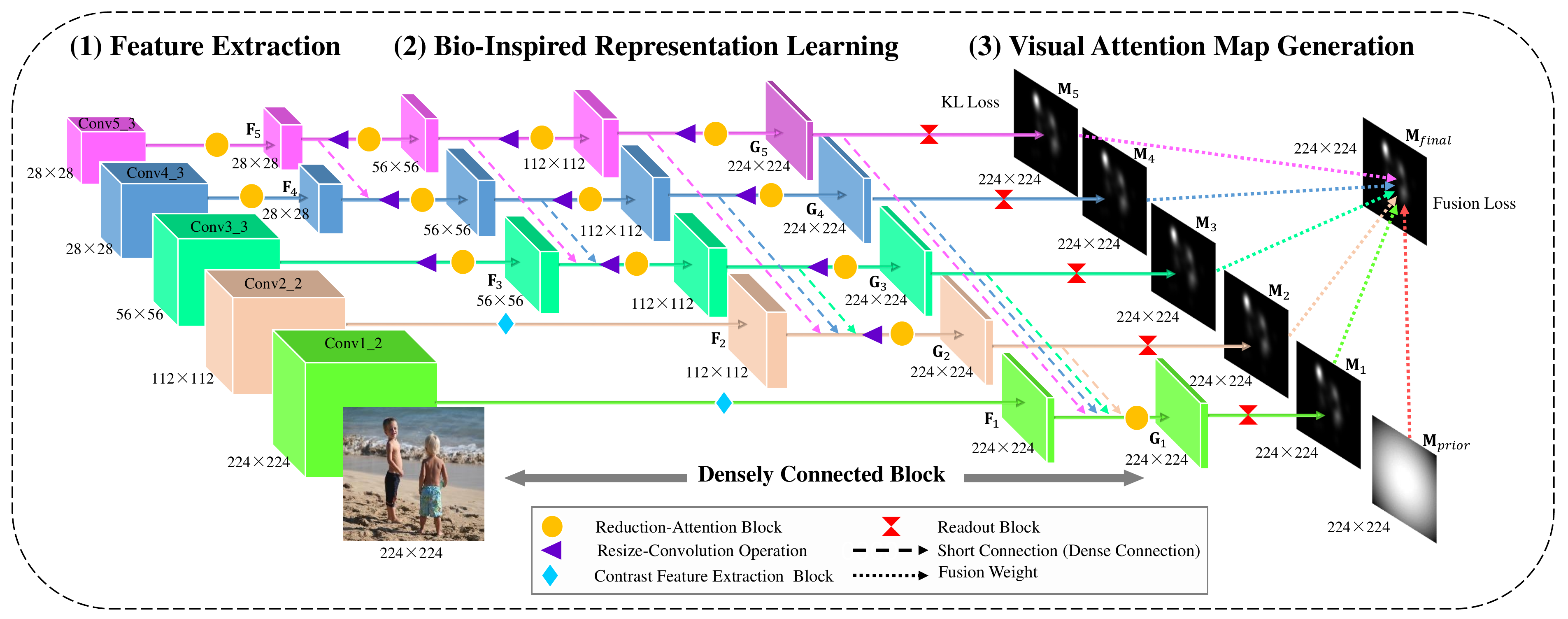}
\renewcommand{\figurename}{Fig.}
\vspace{-8 mm}
\end{center}
    \caption{\small{The network architecture of the proposed method. Firstly, the high-level semantic features are extracted from the refined VGG16, while the low-level contrast features are extracted by the proposed contrast feature extraction block in a deep network. Secondly, during bio-inspired representation learning, both the extracted low-level contrast and high-level semantic features are combined by the designed densely connected block, which is proposed to concatenate various features scale by scale. Finally, the weighted-fusion layer is exploited to generate the ultimate visual attention map based on the obtained representations after bio-inspired representation learning.}}
\label{fig:2}
\end{figure*}

Based on the above facts, an intuitive method is proposed to predict the visual attention map. The proposed method generates the visual attention map by bio-inspired representation learning which can combine both low-level contrast and high-level semantic features. The generated representation coincides with the human visual attention mechanism, since human is sensitive to the patches with high contrast and objects with high semantics. As shown in Fig. \ref{fig:2}, the proposed method can be divided into three main steps: 1) feature extraction, 2) bio-inspired representation learning and 3) visual attention map generation.
Firstly, five different features are extracted from each convolutional block of the refined VGG16 (see \ref{RefinedVGG16}). The features from the first two convolutional blocks may contain the high-frequency contour information, and the features from the last three convolutional blocks may contain information of the internal semantic aspects \cite{Wang2018Deep}. To obtain the low-level contrast features, a contrast feature extraction block is proposed to process the features from the first two convolutional blocks (see \ref{ContrastFeature}) in a deep network. The features from the last three convolutional blocks are usually treated as high-level semantic features.
Secondly, during bio-inspired representation learning, both the extracted low-level contrast and high-level semantic features are combined by the designed densely connected block (see \ref{DenseConnectedBlock}), which is proposed to concatenate various features scale by scale. Specifically, the designed densely connected block introduces some short connections \cite{HouPami18Dss} from higher-level features (deriving from the higher layers of the refined VGG16) to the lower-level features (deriving from the lower layers of the refined VGG16). In this top-down view, the high-level features can provide better locations about the semantic regions, while the low-level features can capture the discriminative regions in the image.
Finally, to generate the ultimate visual attention map, a weighted-fusion layer is utilized to fuse the learned centre-bias prior map and five rough visual attention maps \cite{Zhangtgrs16}. The centre-bias prior map is produced using a two-dimensional Gaussian function, and the five rough visual attention maps are produced by the readout network based on the obtained representations after bio-inspired representation learning (see Fig. \ref{fig:2}). To sum up, the main contributions of this paper are threefold:
\begin{itemize}

  \item A novel network is proposed to learn the bio-inspired representation for visual attention prediction, which can combine both low-level contrast and high-level semantic features.
  \item A contrast feature extraction block is proposed to extract the low-level contrast features in the network. This is the first method applied to extracting low-level contrast features automatically in a deep network.
  \item A densely connected block is proposed by introducing short connections from the higher-level features to the lower-level features. In this top-down view, the high-level features can provide better locations about the semantic regions, while the low-level features can capture rich spatial information.
\end{itemize}

The remaining parts of this paper are organized as follows: Section \ref{relatedworks} reviews the related works. Section \ref{TheProposedmethod} gives a detailed description of the proposed method. The experiments are shown in Section \ref{experiments}. Finally, a conclusion is presented in Section \ref{conclutions}.

\section{Related Works}\label{relatedworks}
In recent years, a large number of visual attention prediction methods have been proposed. According to the kind of the adopted feature, these VAP methods can be mainly divided: 1) methods based on low-level contrast features and 2) methods based on high-level semantic features.

Most of VAP computational methods are based on low-level contrast features. Beginning with the seminal work of Itti {\it et al.} \cite{Itti2002A}, many methods have been proposed for VAP from various mechanisms. On the basis of the mechanism for VAP, existing traditional methods can be classified as: cognitive method \cite{Meur2006A}, information theoretic method \cite{Bruce2010Attention}, graphical method \cite{Sch2006Graph}, spectral analysis method \cite{Hou2007Saliency}, decision theoretic method \cite{Gao2009Decision}, pattern classification method \cite{Judd2010Learning}, and some other methods \cite{Goferman2010Context}. On the one hand, these methods only employ low-level contrast features and do not take into account the semantic aspects (face, people, text, {\it etc.}) which can provide better locations about the semantic regions. On the other hand, the extracted low-level contrast features are insufficient to handle large-scale data with complex distributions because they are hand-crafted.

With the popularity of deep learning, high-level semantic features can be automatically extracted \cite{Hu2016Multimodal, Zhang2013Tensor, Zhang2015Ensemble}, and these high-level semantic features have been exploited for VAP \cite{Vig2014Large, Kmmerer2014b, Kummerer2017Understanding, Cornia2016Predicting, Sss2017DeepFix, Wang2018Deep, Liang2015Predicting, Cong2018Co, Qiu2018Eye, Cong2017An, Han2018CNNs, Wang2018Saliency}. Compared with these methods based on low-level contrast features, methods based on high-level semantic features achieved better results. The first attempt for VAP with deep neural networks was the {\it Ensemble of Deep Networks} (eDN) \cite{Vig2014Large}. However, this method cannot outperform the state-of-the-arts at that time because of its limited training data and poor semantic information for only three convolutional layers. To address the problem in eDN, K$\ddot{\rm u}$mmerer {\it et al.} proposed two methods named Deep Gaze \RNum{1} \cite{K¨¹mmerer2014b} and Deep Gaze \RNum{2} \cite{Kummerer2017Understanding}, and achieved a goodish result. Subsequently, many researchers began to notice the impact of the centre-bias prior on prediction results and a large number of related works emerged. \cite{Cornia2016Predicting, Sss2017DeepFix, Kummerer2017Understanding} introduced centre-bias prior into the deep computational methods for VAP and improved the predicted results to some extent. For the past two years, some researchers on object detection have found that multi-scale features contribute to achieving a good performance for detecting objects of different scales \cite{Wang2018Deep}. Drawing on this success, many researchers adopted multi-scale architecture for VAP, and the predicting performance is further improved by a substantial margin \cite{Wang2018Deep, Liang2015Predicting}. To sum up, most of existing deep learning methods leveraged multi-scale features, introduced centre-bias prior and acquired great success in VAP. However, the utilized features in these methods were not enough to reflect the human visual attention mechanism. Specifically, almost all existing methods based on high-level semantic features overlooked the contribution of low-level contrast features in VAP, which play an important role in visual attention cognition.

Therefore, in order to fully leverage the information of both low-level contrast and high-level semantic features simultaneously, an intuitive VAP method is proposed for generating visual attention map via bio-inspired representation learning. Similar to \cite{HouPami18Dss}, the proposed method introduces some short connections from higher-level features to the lower-level features. But unlike \cite{HouPami18Dss}, we adopt a new processing strategy for the obtained lower-level features because of the task difference. Specifically, VAP aims to locate where the salient objects are, which does not need to highlight the boundaries of the salient objects \cite{HouPami18Dss}. Hence, instead of using the lower-level features directly, we input them into the contrast feature extraction block for automatically obtaining the low-level contrast features in the network. The details of the proposed method will be described in the following sections.

\section{The Proposed method}\label{TheProposedmethod}

In this work, an intuitive method is proposed to generate visual attention map via bio-inspired representation learning. The proposed method is dwelled on in this section. As is depicted in Fig. \ref{fig:2}, the proposed method takes a batch of images with spatial size of 224$\times$224 as input. Firstly, the high-level semantic features $\left\{{\bf{F}}_{3},{\bf{F}}_{4},{\bf{F}}_{5}\right\}$ are extracted from the refined VGG16 (see \ref{RefinedVGG16}), while the low-level contrast features $\left\{{\bf{F}}_{1}, {\bf{F}}_{2}\right\}$ are extracted by the proposed contrast feature extraction block (see \ref{ContrastFeature}) in a deep network.
Secondly, during bio-inspired representation learning, both the extracted low-level contrast and high-level semantic features are combined by the designed densely connected block (see \ref{DenseConnectedBlock}), which is proposed to concatenate various features scale by scale.
Finally, the weighted-fusion layer is exploited to generate the ultimate visual attention map ${\bf{M}}_{final}$ based on the obtained representations $\left\{{\bf{G}}_{1},{\bf{G}}_{2},{\bf{G}}_{3},{\bf{G}}_{4},{\bf{G}}_{5}\right\}$ after bio-inspired representation learning.
The details are introduced as follows.

\subsection{Feature Extraction}\label{featureExtraction}
This step can be divided into two parts: the refined VGG16 and contrast feature extraction block. Thereinto, the first part is used for extracting high-level semantic features (the outputs of $conv$3-3, $conv$4-3 and $conv$5-3 in the refined VGG16) and raw low-level features ($conv$1-2 and $conv$2-2 in the refined VGG16). The second part is used for obtaining low-level contrast features from the raw low-level features, which can extract low-level contrast features in a deep network. The refined VGG16 and contrast feature extraction block are elaborated as follows.
\subsubsection{\textbf{The Refined VGG16}}\label{RefinedVGG16}
The CNNs require the size of the input image to be fixed. For example, existing pre-trained VGG models \cite{Simonyan15} require the input size to be 224$\times$224. This requirement is determined by the network structure, because CNNs generally include multiple fully connected layers, and the number of fully connected neurons is usually fixed. Generally, multi-level features (the output of different layer in the neural network) can effectively improve the results of image retrieval, image classification, and object detection tasks \cite{Zhao2018Embedding}. The same is true for the VAP task. Differently, the VAP task only needs to locate the most salient region and has nothing to do with the object boundary. Therefore, we extract the low-level features for refining contrast features instead of improving the edge extraction result. To obtain multi-level features, we use a strategy of feature pyramid network to extract various features from different layers.

In this paper, we refine VGG16 for extracting multiple features on account of its elegance, simplicity and effectiveness on extracting various visual features. The origin VGG16 consists of five convolutional blocks, where each block is followed by a max-pooling layer, and three fully connected layers. Considering what we want to harness is the feature map, the origin VGG16 network is refined. We only adopt the convolutional blocks except for the last pooling layer and omit fully connected layers. In addition, to avoid the feature map being too small in the fifth convolutional block, we set the stride of the fourth pooling as 1, and meantime use a convolutional kernel with holes of size 2 in the fifth convolutional block to keep the same receptive filed as the origin architecture.

As is shown in Fig. \ref{fig:2}, formulaically, let ${\bf{I}}\in \mathbb{R}^{224 \times 224 \times 3}$ represent one of the input images. For the $m$-th convolutional layer, we denote its input as ${\bf{I}}_{m}$, where $m\in$\{$1,\cdots,13$\}, and its output as ${\bf{X}}_{m}$. As for the input of each convolutional layer, a kernel ${\bf{W}}_{m}$ is adopted to perform 2D convolution operation. And then a bias term ${\bf{b}}_{m}$ is added to the convolution operation result. An activation function is used for non-linear mapping. The process of each convolutional layer can be formulated as:
\begin{equation}\label{Eq.:1}
{\bf{X}}_{m}=\delta\left({\bf{I}}_{m} \ast {\bf{W}}_{m}+{\bf{b}}_{m}\right),
\end{equation}
where $\ast$ denotes convolution operation and $\delta\left(\cdot\right)$ indicates activation function. The {\it Rectified Linear Unit} (ReLU)  is adopted as activation function in this paper for its high efficiency and it can be formulated as:
\begin{equation}\label{Eq.:2}
\delta\left(x\right)=max\left(0,x\right),
\end{equation}
where $x$ denotes an arbitrary real number. In addition, each convolutional block is followed by a max-pooling layer. For these convolutional blocks, the weights of filters in the first four blocks are initialized from VGG16 network. Five different features are obtained from the outputs of $conv$1-2, $conv$2-2, $conv$3-3, $conv$4-3 and $conv$5-3. Then, the outputs of $conv$3-3, $conv$4-3 and $conv$5-3 are directly input into the bio-inspired representation learning step, and the outputs of $conv$1-2 and $conv$2-2 are fed into the contrast feature extraction block for automatically obtaining contrast features.

\subsubsection{\textbf{Contrast Feature Extraction Block}}\label{ContrastFeature}

\begin{figure}[tp]
\begin{center}
\includegraphics[width=0.7\linewidth]{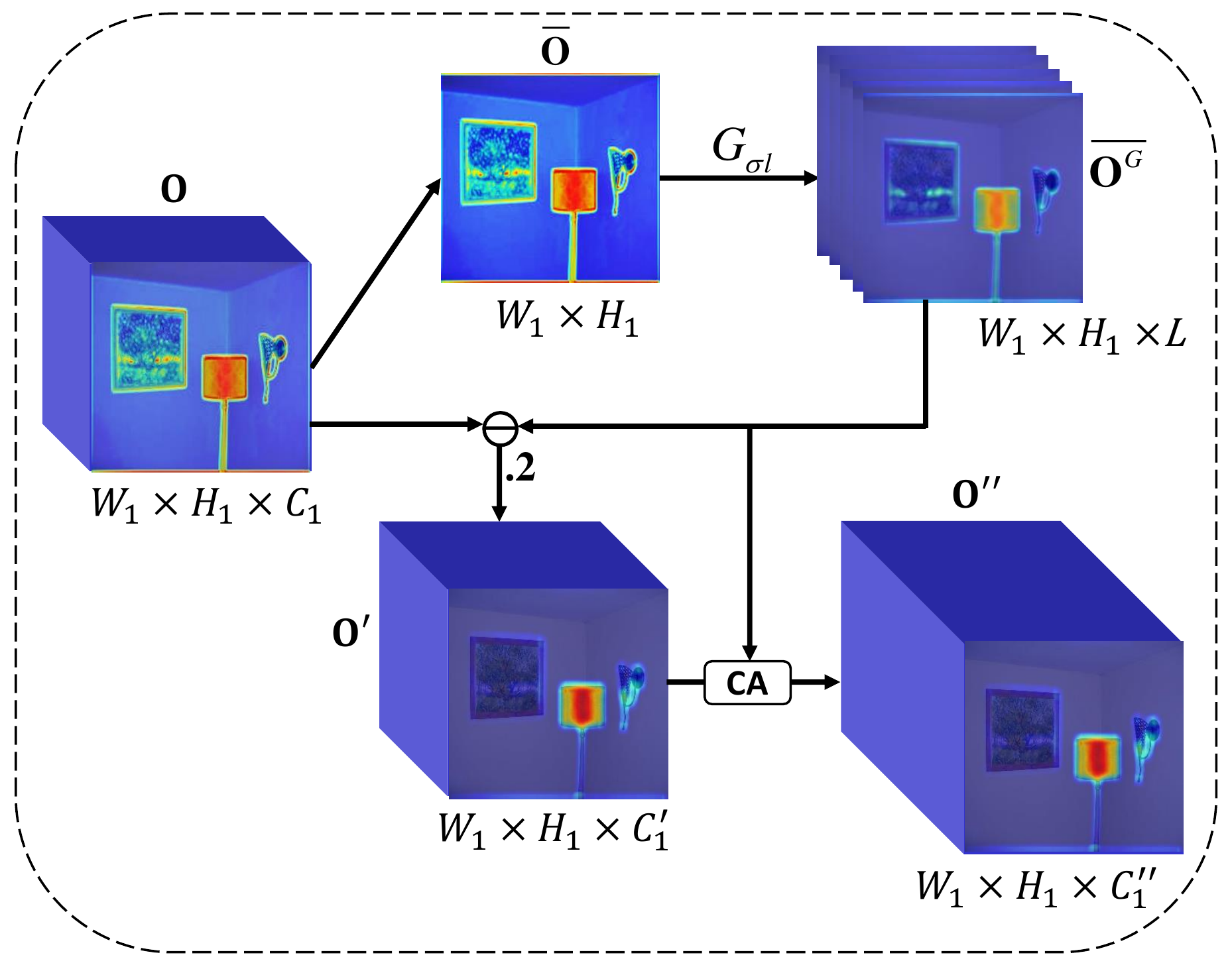}
\renewcommand{\figurename}{Fig.}
\vspace{-5 mm}
\end{center}
    \caption{\small{The devised contrast feature extraction block. Firstly, we take the outputs of $conv$1-2 and $conv$2-2 in the refined VGG16 as the inputs. Secondly, the input features are processed into a Gaussian Pyramid by five different Gaussian kernels. Thirdly, we compute the squared residual features by subtracting each channel of the input features from the Gaussian Pyramid and then square them. Finally, eventual contrast features are obtained via adaptively merging squared residual features and Gaussian Pyramid features.}}
\label{fig:3}
\end{figure}

Considering that human eyes are sensitive to high contrast patches, a contrast feature extraction block (Fig. \ref{fig:3}) is devised for low-level contrast features automatical extraction in a deep network. The outputs of $conv$1-2 and $conv$2-2 in the refined VGG16 may contain the high-frequency contour information, which is the edge of high contrast patches. As a result, we take the outputs of $conv$1-2 and $conv$2-2 in the refined VGG16 as the inputs of this contrast feature extraction block for automatically obtaining contrast features in deep network. Let ${\bf{O}}=\left[{\bf{o}}_{1},{\bf{o}}_{2},\cdots,{\bf{o}}_{C_{1}}\right]\in \mathbb{R}^{W_{1} \times H_{1} \times C_{1}}$ denote the input of this block. Firstly, the input ${\bf{O}}$ is squeezed as one channel in an average manner across all channels. We call this one-channel feature map as intensity map, which is denoted as $\overline{{\bf{O}}}\in \mathbb{R}^{W_{1} \times H_{1}}$. Secondly, five different Gaussian kernels are employed for ${\bf{O}}$ to get a Gaussian Pyramid $\overline{{\bf{O}}^{G}}=\left[\overline{{\bf{o}}^{G}_{1}},\overline{{\bf{o}}^{G}_{2}},\cdots,\overline{{\bf{o}}^{G}_{L}}\right]\in \mathbb{R}^{W_{1} \times H_{1}\times L}$ \cite{Kummerer2017Understanding}, where $L$=5. The obtained Gaussian Pyramid by different Gaussian kernels contains different spatial information with various scale. Formulaically,
\begin{equation}\label{Eq.:3}
\overline{{\bf{o}}^{G}_{l}}=\overline{{\bf{O}}}\ast G_{\sigma l},
\end{equation}
where the standard deviations the Gaussian kernels $G_{\sigma l}$ are 5, 10, 20, 40 and 80 pixels, respectively. Thirdly, the residual features ${\bf{O}}^{'}\in \mathbb{R}^{W_{1} \times H_{1} \times C_{1}^{'}}$ are computed by subtracting each channel of the input features ${\bf{O}}$ from the Gaussian Pyramid $\overline{{\bf{O}}^{G}}$, where $C_{1}^{'}$=5$\times C_{1}$. Fourthly, we get squared residual features by squaring these residual features. Finally, eventual contrast features are obtained via adaptively merging squared residual features and Gaussian Pyramid features instead of simply concatenating them. Let ${\bf{O}}^{''}\in \mathbb{R}^{W_{1} \times H_{1} \times C_{1}^{''}}$ represent the eventual contrast features, then the whole process can be expressed as the following formulation:
\begin{equation}\label{Eq.:4}
\begin{split}
{\bf{O}}^{''}&=\mathcal{CF}(\bf{O})=\mathcal{AM}({\bf{O}}^{'},\overline{{\bf{O}}^{G}})\\
&=\left(\sum_{c_{1}=1}^{C_{1}}\sum_{l=1}^{L}({\bf{o}}_{c_{1}}-\overline{{\bf{o}}^{G}_{l}})^{2}\right)\ast {\bf{W}}_{AM}+{\bf{\overline{{\bf{O}}^{G}}}}\ast{\bf{W}}_{AM}^{'},
\end{split}
\end{equation}
where $\mathcal{CF}(\cdot)$ denotes the whole contrast feature extraction operation, $\mathcal{AM}(\cdot,\cdot)$ indicates the adaptively merging operation, ${\bf{W}}_{AM}$ and ${\bf{W}}_{AM}^{'}$ are the adaptively merging weights.

To validate the effectiveness of the proposed contrast feature extraction block, the comparative experiment is performed and the experiment result is shown in Table \ref{tab:5}.

\subsection{Bio-Inspired Representation Learning}
The insight of the proposed method is that both low-level contrast and high-level semantic features contribute to VAP. As a result, both low-level contrast and high-level semantic features should be taken into account. To this end, bio-inspired representation learning step is devised. This step can be divided into two main parts: 1) the reduction-attention block for alleviating the problem of dimensionality disaster and adaptively recalibrating channel-wise feature responses while combining various features during bio-inspired representation learning, and 2) the densely connected block for combining both low-level contrast and high-level semantic features by a concatenation operator scale by scale. We elaborate these two parts hereinafter.

\subsubsection{\textbf{Reduction-Attention Block}}
Traditional convolutional neural network is a kind of spatial attention network by learning different filters. Therefore, it only considers the local connectivity pattern on each input channel, and the input channels are treated with a uniform weight. Recent works \cite{hu2018senet, Chen2017SCA, Wang2018Resilient} have demonstrated that different channels of the feature map contain different amount of information. In other words, different channels of the feature map should be allocated with suitable weights which match their amount of information. Accordingly, inspired by the SE block \cite{hu2018senet}, a new reduction-attention block is devised. The devised reduction-attention block can process the original feature into a channel descriptor, which contains global information about each channel feature and can discard some interferences in the intermediate representation. In general, the devised reduction-attention block aims to avoid overfitting and learn a series of weights for each input channel. Specifically, to ease the problem of dimensionality disaster while performing bio-inspired representation learning \cite{Yao2017Revisiting}, a dimension reduction architecture is used for obtaining the reductive feature in the first place. Then, a retrofitted SE block is followed by the adaptive dimension reduction architecture. Different from the origin SE block, the second fully connected layer is omitted and the input is reductive in the retrofitted SE block to alleviate overfitting.

\begin{figure}[tp]
\begin{center}
\includegraphics[width=\linewidth]{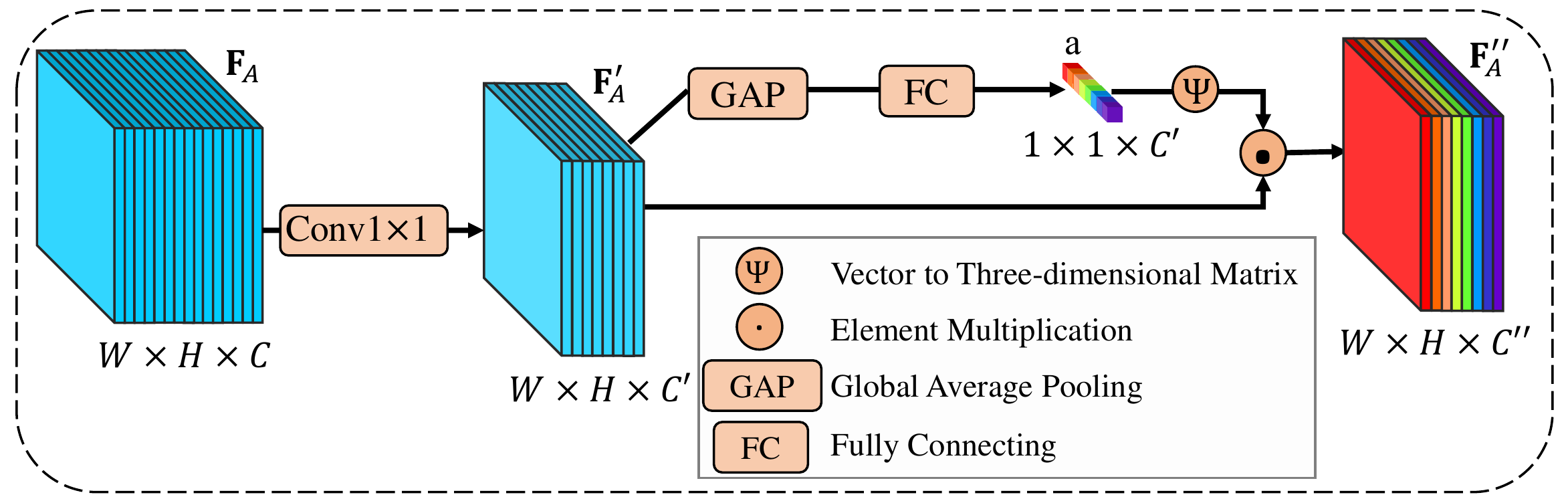}
\renewcommand{\figurename}{Fig.}
\vspace{-7 mm}
\end{center}
    \caption{\small{The devised reduction-attention block. Firstly, the input feature is reduced by a $1 \times 1$ convolution layer. Secondly, the reduced feature is squeezed to get a channel descriptor. Thirdly, the obtained channel descriptor is extended to a channel-weighted matrix. And finally, the channel-weighted feature is obtained by element-wise multiplication between the input feature and the channel-weighted matrix.}}
\label{fig:4}
\end{figure}

As is shown in Fig. \ref{fig:4}, firstly, the feature dimension of ${\bf{F}}_{A}\in \mathbb{R}^{W \times H \times C}$ is reduced by a operation of 1$\times$1 convolutional layer to get a more elegant feature ${\bf{F}}^{'}_{A}\in \mathbb{R}^{W \times H \times C'}=[{\bf{f}}^{'}_{1},{\bf{f}}^{'}_{2},\cdots,{\bf{f}}^{'}_{C'}]$. Meanwhile, the nonlinear interaction between channels is increased after the 1$\times$1 convolutional layer. Secondly, the global spatial information of each channel in ${\bf{F}}^{'}_{A}$ is squeezed so as to generate a channel descriptor. Thirdly, in order to convert the generated channel descriptor into the channel weight ${\bf{a}}=\left[a_{1},a_{2},\cdots,a_{C}\right] \in \mathbb{R}^{C}$, channel-wise dependencies need to be fully captured, and a non-mutually-exclusive relationship is supposed to learn. To this end, a fully connected layer and a sigmoid activation are employed to the generated channel descriptor \cite{hu2018senet}. The $c$-th element of $\bf{a}$ can be calculated by:
\begin{equation}
a_{c}=\sigma\left(\frac{1}{W\times H}\sum^{W}_{s=1}\sum^{H}_{k=1}{\bf{f}}^{'}_{c}\left(s,k\right)\right),
\end{equation}\label{Eq.:5}
where ${\bf{f}}^{'}_{c}\left(s,k\right)$ indicates the response of ${\bf{f}}^{'}_{c}$ at location $(s,k)$, and $\sigma$ represents sigmoid activation function, which can be formulated as:
\begin{equation}\label{Eq.:6}
\sigma(x)=\frac{1}{1+e^{-x}},
\end{equation}
where $x$ denotes an arbitrary real number. Finally, the channel-weighted feature ${\bf{F}}^{''}_{A}$ is acquired by the following equation:
\begin{equation}\label{Eq.:7}
{\bf{F}}^{''}_{A}=\hbar({\bf{F}}_{A})=\Psi(\bf{a})\odot {\bf{F}}^{'}_{A}={\bf{\overline W}}\odot{\bf{F}}^{'}_{A},
\end{equation}

where $\hbar(\cdot)$ indicates the reduction-attention operation, $\odot$ represents element-wise multiplication, and $\Psi(\cdot)$ denotes the operation of extending a vector to a three-dimensional matrix, which is defined by us. Specifically for $\Psi(\cdot)$, a channel weight matrix ${\bf{\overline W}}=\left[{\bf{\overline w}}_{1},{\bf{\overline w}}_{2},\cdots,{\bf{\overline w}}_{C}\right]\in \mathbb{R}^{W \times H \times C}$ is generated, in which all values of ${\bf{\overline w}}_{c}\in \mathbb{R}^{W \times H}$ are equal to the value of $a_{c}$.

To validate the effectiveness of the proposed reduction-attention block, the comparative experiment is performed and the experiment result is shown in Table \ref{tab:5}.

\subsubsection{\textbf{Densely Connected Block}} \label{DenseConnectedBlock}
As is shown in Fig. \ref{fig:2}, the densely connected block is based on DenseNet \cite{Huang2017Densely} and the work of Cheng {\it et al.} \cite{HouPami18Dss}. We choose five-level features $\left\{{\bf{F}}_{1},{\bf{F}}_{2},{\bf{F}}_{3},{\bf{F}}_{4},{\bf{F}}_{5}\right\}$. Thereinto, ${\bf{F}}_{3}$, ${\bf{F}}_{4}$ and ${\bf{F}}_{5}$ are treated as high-level semantic features, which directly derived from the refined VGG16, while ${\bf{F}}_{1}$ and ${\bf{F}}_{2}$ are considered as low-level contrast features, which come from the contrast feature extraction block. In this densely connected block, five-level ultimate feature representations $\left\{{\bf{G}}_{1},{\bf{G}}_{2},{\bf{G}}_{3},{\bf{G}}_{4},{\bf{G}}_{5}\right\}$ are obtained by lateral connecting scale by scale and cross-level dense connecting. For instance, in order to get ${\bf{G}}_{1}$, the channel numbers of features ${\bf{F}}_{1},{\bf{F}}_{2},{\bf{F}}_{3},{\bf{F}}_{4}$ and ${\bf{F}}_{5}$ are reduced using the designed reduction-attention block. Then, the resize-convolution operation \footnote{https://distill.pub/2016/deconv-checkerboard/} is introduced for upsampling all the obtained multiple features except for ${\bf{F}}_{1}$. The reason for choosing the resize-convolution operation is its advantage in averting the checkerboard effect. Finally, ${\bf{F}}_{1}$ and other upsampling features are combined to get the ultimate representation ${\bf{G}}_{1}$. After the iteration above, the other four ultimate representations ${\bf{G}}_{2}$, ${\bf{G}}_{3}$, ${\bf{G}}_{4}$, and ${\bf{G}}_{5}$ are acquired. Here, we suppose that there are total $J$ ultimate representations or $J$ branches. Formulaically,

\begin{equation}\label{Eq.:8}
{\bf{G}}_{j}=
\left\{
\begin{array}{lr}
\mathcal{A}^{[j-2]}\left({\bf{F}}_{j}\right),&if\ j=J\\
\\
\hbar\left(\sum\limits_{i=j}^{J-1}{\bf{w}}^{j}_{i}{\bf{R}}^{i-j}_{i}+{\bf{R}}^{J-1-j}_{J}\right), &else\ if\ j=1\\
\\
\mathcal{A}^{[j-1]}\left(\sum\limits_{i=j}^{J-1}{\bf{w}}^{j}_{i}{\bf{R}}^{i-j}_{i}+{\bf{R}}^{J-1-j}_{J}\right),&otherwise\\
\end{array}
\right.
\end{equation}
where $\mathcal{A}^{[j-2]}(\cdot)$ and $\mathcal{A}^{[j-1]}(\cdot)$ denote the tandem operation of resize-convolution operation and reduction-attention operation, in which the superscript $j$-$2$ and $j$-$1$ signify the count of this kind of operation. ${\bf{R}}^{i-j}_{i}$ denotes the activation of ${\bf{F}}_{i}$ after $\mathcal{A}^{[i-j]}(\cdot)$, and similarly, ${\bf{R}}^{J-1-j}_{i}$ denotes the activation of ${\bf{F}}_{i}$ after $\mathcal{A}^{[J-1-j]}(\cdot)$. ${\bf{w}}^{j}_{i}$ denotes the weight of the short connection from the $i$-th branch to the $j$-th branch. Note that $\mathcal{A}^{[0]}(\cdot)=1$ and ${\bf{R}}^{0}_{j}={\bf{F}}_{j}$.

To validate the effectiveness of the devised densely connected block, the comparative experiment is performed and the experiment result is shown in Table \ref{tab:5}.

\subsection{Visual Attention Map Generation}
In this step, we aim to predict the final visual attention map based on the obtained features after bio-inspired representation learning. Firstly, five rough visual attention maps are generated by the readout network, which is stacked by three successive reduction-attention blocks. And subsequently, a weighted-fusion layer is adopted for fusing these five rough visual attention maps with the learned centre-bias prior map to generate the final visual attention map. The learned centre-bias prior map is added to compensate for the secondary effect by the photographer¡¯s bias. Hereinafter, we dwell on this step in two main parts: centre-bias prior block and loss function.

\subsubsection{\textbf{Centre-bias Prior Block}}
Physiological studies have shown that when observing an image, human attention tends to concentrate on the centre. This is a common phenomenon named centre-bias, which is because the photographer is accustomed to placing the object of interest at the centre of the image. As a result, when people watch such images repeatedly, their attention is attracted by the centre of the image naturally. Note that when there are no highly salient objects, human attention is also attracted to the centre of the image. Therefore, this phenomenon has great influence on the final evaluation metrics, such as {\it Area Under Curve} (AUC), {\it Normalized Scanpath Saliency} (NSS), {\it Linear Correlation Coefficient} (CC) and so on, even leading to unfair evaluation.

In order to address this problem, a learnable centre-bias prior block is designed using a two-dimensional Gaussian function for modeling the centre-bias. Compared to most of existing works, the proposed centre-bias prior is learnable and derives from data purely instead of the assumptions from indistinct biological studies. The insight of our centre-bias prior block comes from the work of Conia {\it et al.} \cite{Cornia2016Predicting}, which also adopts a learnable manner modeling the centre-bias prior. Differently, our centre-bias prior block is more effective with less parameter than theirs. Specifically, a specifical Gaussian map is generated as Eq. \ref{Eq.:9} depending on the learnable specifical horizontal variance $\sigma^{2}_{x}$ and vertical variance $\sigma^{2}_{y}$. Nevertheless, the method of Conia {\it et al.} \cite{Cornia2016Predicting} relies on learnable means and variances, in which there is no need to learn means since the center of the generated Gaussian probability distribution is fixed. The equation of generating centre-bias prior map is as follows:
\begin{equation}\label{Eq.:9}
{\bf{M}}_{prior}=\frac{1}{2\pi\sigma_{x}\sigma_{y}}{\rm exp}\left(-\left(\frac{(x-x_{0})^2}{2\sigma^2_{x}}+\frac{(y-y_{0})^2}{2\sigma^2_{y}}\right)\right).
\end{equation}
Here, each Gaussian map represents a spatial pattern. We can model the centre-bias using Gaussian map purely from data rather than the assumptions from indistinct biological studies.

To validate the effectiveness of the proposed centre-bias prior block, the comparative experiment is performed and the experiment result is shown in Table \ref{tab:5}.

\subsubsection{\textbf{Loss Function}}
To get five rough visual attention maps, the obtained five features ${{\bf{G}}_{1}}$, ${{\bf{G}}_{2}}$, ${{\bf{G}}_{3}}$, ${{\bf{G}}_{4}}$, and ${{\bf{G}}_{5}}$ after the inspired representation learning are fed into the readout network. The readout network is stacked by three reduction-attention blocks to learn interactions across channels instead of learning new spatial features. Let ${\bf{M}}_{j}$ represent the obtained $j$-th rough visual attention map, and ${\bf{W}}_{j}$ denotes all the parameters before generating the $j$-th rough visual attention map ${\bf{M}}_{j}$. For simplicity, the bias term is absorbed into ${\bf{W}}_{j}$. Therefore, the objective function for generating the $j$-th rough visual attention map can be given with Kullback-Leibler divergence and weight decay term as follows:
\begin{equation}\label{Eq.:10}
\begin{split}
&L_{j}\left({\bf{M}}_{j},{\bf{Z}}_{den}\right)={\rm KL}\left({\bf{M}}_{j},{\bf{Z}}_{den}\right)+\alpha\sum^{T}_{t}\left({\rm {W}}_{j}^{t}\right)^{2}\\
&=\sum^{T}_{t}{\bf{Z}}_{den}^{\left(t\right)}{\rm log}\left(\frac{{\bf{Z}}_{den}^{\left(t\right)}}{{\bf{M}}^{\left(t\right)}_{j}+\epsilon}+\epsilon\right)+\alpha\sum^{T}_{t}\left({\rm {W}}_{j}^{(t)}\right)^{2},
\end{split}
\end{equation}
where ${\rm KL}\left(\cdot,\cdot\right)$ means the Kullback-Leibler divergence of two probability distributions, ${\bf{Z}}_{den}$ represents the groundtruth density map, $t$ indexes the $t$-th pixel, $\epsilon$ denotes the regularization constant to avoid the Nan value of the loss during training phase, $\alpha$ is the coefficient of weight penalty, and ${\rm {W}}_{j}^{(t)}$ indicates the value of the $t$-th pixel in matrix ${\bf{W}}_{j}$.

Next, to take advantage of the merit of different rough visual attention maps and tackle the secondary effect by the photographers bias, a weighted-fusion layer is added for image-level fusion. Note that the architecture of the weighted-fusion layer is similar to the readout network. The weighted-fusion layer is also stacked by three reduction-attention blocks to learn interactions across different rough visual attention maps and the learned centre-bias prior map. Finally, after the weighted-fusion layer, a Gaussian convolutional layer is adopted to produce the final visual attention map. As a result, when we suppose that the parameter in the weighted-fusion layer is $\bf{K}$, the final visual attention map can be given as follows:
\begin{equation}\label{Eq.:11}
{\bf{M}}_{final}=\left(\sum_{j=1}^{J}{\bf{M}}_{j}+{\bf{M}}_{prior}\right)\ast {\bf{K}} \ast G_{\sigma},
\end{equation}
where ${\bf{M}}_{final}$ is the final visual attention map, and $G_{\sigma}$ represents the Gaussian convolution kernel. Therefore, the final fusion loss function can be formulated as:
\begin{equation}\label{Eq.:12}
L_{fuse}\left({\bf{M}}_{final},{\bf{Z}}_{den}\right)={\rm KL}\left({\bf{M}}_{final},{\bf{Z}}_{den}\right).
\end{equation}
Ultimately, the final loss function can be calculated by:
\begin{equation}\label{Eq.:13}
\begin{split}
L_{final}({\bf{M}}_{j},{\bf{M}}_{final},{\bf{Z}}_{den})&=\sum_{j=1}^{M}L_{j}({\bf{M}}_{j},{\bf{Z}}_{den})\\
&+L_{fuse}\left({\bf{M}}_{final},{\bf{Z}}_{den}\right).
\end{split}
\end{equation}
Then all parameters ${\bf{W}}_{j}$ and ${\bf{K}}$ can be learned by minimizing the final loss $L_{final}({\bf{M}}_{j},{\bf{M}}_{final},{\bf{Z}}_{den})$ over all the training images via RMSProp. After training, given a test image ${\bf{I}}_{test}$, a corresponding visual attention map ${\bf{M}}_{test}$ can be predicted by the proposed method. The main procedure of the proposed method is shown in Algorithm \ref{alg:Framwork}.

\begin{algorithm}[htb]
\renewcommand{\algorithmicrequire}{\textbf{Input:}}
\renewcommand\algorithmicensure {\textbf{Output:} }

\caption{The proposed method}
\label{alg:Framwork}
\begin{algorithmic}[1]

\REQUIRE ~~\\

Training images $\bf{I}$ and their corresponding labels ${\bf{Z}}_{den}$,\\
Testing image ${\bf{I}}_{test}$.

\ENSURE ~~\\
Testing visual attention map ${\bf{M}}_{test}$,\\
All the parameters ${\bf{W}}_{j}$ of generating the $j$-th visual attention map, \\
All the parameters $\bf{K}$ in the weighted-fusion layer.

\renewcommand{\algorithmicrequire}{\textbf{Initialization:}}
\REQUIRE ~~\\
The weights of the first four convolutional blocks in the refined VGG16 are initialized from the origin VGG16, and the remaining weights are randomly sampled by truncated\_normal distribution.
\renewcommand{\algorithmicrequire}{\textbf{Repeat:}}
\REQUIRE ~~\\

\STATE Calculate the feature map ${\bf{F}}_{j}$;

\STATE Learn the bio-inspired feature representations ${\bf{G}}_{j}$;

\STATE Infer the $j$-th rough visual attention map ${\bf{M}}_{j}$ and calculate the loss $L_{j}$ according to Eq. \ref{Eq.:10};

\STATE Generate the centre-bias prior map ${\bf{M}}_{prior}$ according to Eq. \ref{Eq.:9};

\STATE Infer the final visual attention map ${\bf{M}}_{final}$ and calculate the fusion loss $L_{fuse}$ according to Eq. \ref{Eq.:12};

\STATE Calculate the final loss $L_{final}$ according to Eq. \ref{Eq.:13};

\STATE Update the parameters ${\bf{W}}_{j}$ and $\bf{K}$ by utilizing RMSProp.

\renewcommand\algorithmicensure {\textbf{Until:}{\,}{A fixed number of iterations}}
\ENSURE ~~\\
\STATE Generate the testing visual attention map ${\bf{M}}_{test}$.
\renewcommand\algorithmicensure {\textbf{Return:}{\,}{${\bf{M}}_{test}$, ${\bf{W}}_{j}$, $\bf{K}$}}
\ENSURE ~~\\
\end{algorithmic}

\end{algorithm}

\section{Experiment and Results}\label{experiments}
In this section, we elaborate datasets, evaluation metrics and implementation details about the proposed method. In addition, we compare our experiment results with state-of-the-arts and give an ablation analysis for each component.

\subsection{Datasets}
The proposed method is trained on SALICON dataset \cite{Jiang2015SALICON}, the largest dataset for visual attention prediction. And it is tested on four challenging saliency datasets, including OSIE \cite{Xu2014Predicting}, MIT1003 \cite{Judd2010Learning}, TORONTO \cite{Bruce:2005} and PASCAL-S datasets \cite{Li2014The}. All the stimulus images in the dataset are common natural images. The groundtruth fixation maps are obtained from fixation locations of observers for the stimulus images. And the groundtruth density maps are generated by convolving the fixation location points in the groundtruth fixation maps using a Gaussian filter. A brief description is given for 5 saliency datasets used in training and testing as follows.

\subsubsection{\textbf{SALICON \cite{Jiang2015SALICON}}} The SALICON dataset is the largest dataset for selective attention prediction. It contains 20000 images, including 10000 for training, 5000 for validating and 5000 for testing. This dataset is based on mouse-contingent-tracking on multi-resolution stimulus images, which are taken from Microsoft COCO dataset \cite{Lin2014Microsoft}. In this work, 10000 training images and 5000 validating images are leveraged.

\subsubsection{\textbf{OSIE \cite{Xu2014Predicting}}} The OSIE dataset contains 700 natural indoor and outdoor scenes, aesthetic photographs from Flickr and Google. Each image corresponds to an eye tracking annotation from 15 viewers.

\subsubsection{\textbf{MIT1003 \cite{Judd2010Learning}}} The MIT1003 dataset includes 1003 natural indoor and outdoor scenes, which are randomly selected from Flickr creative commons and LabelMe. The groundtruth map is generated by recording eye tracking data from 15 viewers when they observe these images freely.

\subsubsection{\textbf{TORONTO \cite{Bruce:2005}}} The TORONTO dataset involves 120 color images of outdoor and indoor scenes. Corresponding fixation maps are generated by the eye tracking data from 20 subjects. This dataset is widly used to evaluate the method of VAP.

\subsubsection{\textbf{PASCAL-S \cite{Li2014The}}} The PASCAL-S dataset is a collection of 850 natural images from the validation set of PASCAL VOC 2010. The groundtruth maps are generated by the eye tracking data during 2 seconds from 8 observers.
\subsection{Evaluation Metrics}
There are several evaluation metrics to measure the consistency between the predicted result and the groundtruth \cite{Riche2014Saliency}. These metrics can be roughly divided into three categories: 1) the distribution-based metric, 2) the value-based metric and 3) the location-based metric. The distribution-based metrics include {\it Earth Movers Distance} (EMD) and {\it Linear Correlation Coefficient} (CC). The value-based metric involves {\it Normalized Scanpath Saliency} (NSS). And the location-based metric is mainly {\it Area Under Curve} (AUC), which consists of several variations: AUC-Judd, AUC-Borji, shuffled-AUC. Here, for clarity, the predicted visual attention map, the groundtruth density map and the groundtruth fixation map are denoted as ${\bf{M}}_{final}$, ${\bf{Z}}_{den}$, and ${\bf{Z}}_{fix}$, respectively. Next, these metrics are elaborated in the following.

\subsubsection{\textbf{Earth Movers Distance (EMD)}} EMD can be used to measure the normalized minimum cost of changing from one distribution to another. Therefore, the performance of the proposed method can be evaluated by calculating the EMD between the predicted visual attention map ${\bf{M}}_{final}$ and the groundtruth density map ${\bf{Z}}_{den}$. The smaller EMD is, the smaller the distance between ${\bf{M}}_{final}$ and ${\bf{Z}}_{den}$ is, and the better performance of the proposed method is.

\subsubsection{\textbf{Linear Correlation Coefficient (CC)}} CC is also called the Pearson linear correlation coefficient. It can be used to measure the linear correlation coefficient between the predicted visual attention map ${\bf{M}}_{final}$ and the groundtruth density map ${\bf{Z}}_{den}$. The closer to 1 the score of CC is, the higher the linear correlation between ${\bf{M}}_{final}$ and ${\bf{Z}}_{den}$ is, and the better performance of the proposed method is. The formula is as follows:
\begin{equation}\label{Eq.:14}
  CC=\frac{cov({\bf{M}}_{final},{\bf{Z}}_{den})}{\sigma^{'}({\bf{M}}_{final})\times\sigma^{'}({\bf{Z}}_{den})},
\end{equation}
where $\sigma^{'}({\bf{M}}_{final})$ and $\sigma^{'}({\bf{Z}}_{den})$ are the standard deviation corresponding to the the predicted visual attention map ${\bf{M}}_{final}$ and the groundtruth density map ${\bf{Z}}_{den}$, and $cov({\bf{M}}_{final},{\bf{Z}}_{den})$ denotes the covariance between ${\bf{M}}_{final}$ and ${\bf{Z}}_{den}$.

\subsubsection{\textbf{Normalized Scanpath Saliency (NSS)}} NSS can be used to measure the saliency value at human fixations. Let ${\bf{M}}^{'}_{final}$ represent the normalized predicted visual attention map. Then NSS can express the average value of ${\bf{M}}^{'}_{final}$ at all human fixations. Given the predicted visual attention map ${\bf{M}}_{final}$ and the groundtruth fixation map ${\bf{Z}}_{fix}$, NSS can be calculated by the following formula:
\begin{equation}\label{Eq.:15}
\begin{split}
&NSS=\frac{1}{T'}\sum_{t=1}^{T'}{\bf{M}}^{'}_{final}(t)\times {\bf{Z}}_{fix}(t)\\
&{\rm{where}}\quad T'=\sum_{t}{\bf{Z}}_{fix}(t),\\
& {\rm{and}} \quad {\bf{M}}^{'}_{final}=\frac{{\bf{M}}_{final}-\mu({\bf{M}}_{final})}{\sigma^{'}({\bf{M}}_{final})}.
\end{split}
\end{equation}

Here, $T'$ represents the number of all fixated pixels. NSS is sensitive to the false positive. Hence, the larger NSS is, the higher the accuracy of the predicted visual attention map is.

\subsubsection{\textbf{Area Under Curve (AUC)}} AUC is a location-based metric. The groundtruth fixation map ${\bf{Z}}_{fix}$ contains human fixation information for the image. According to ${\bf{M}}_{final}$ and ${\bf{Z}}_{fix}$, the true positive rate and the false positive rate can be calculated, and then the receiver operating characteristic (ROC) curve is plotted. By calculating the area under the ROC curve, the AUC score is obtained. Depending on the choice of non-fixation distribution, three different variations of AUC are adopted to evaluate the performance of the proposed method: AUC-Judd, AUC-Borji, and shuffled-AUC (s-AUC). The first two versions choose the non-fixation in the image as a uniform distribution, while the s-AUC uses the human fixation in other images under the dataset as non-fixation distribution. In general, the s-AUC is more reasonable because it gives a penalization to the methods which consider the centre bias in eye fixations \cite{Cornia2016Predicting}. The score of AUC should range 0.5 to 1. And the closer to 1 the score is, the better performance of the proposed method is.

\subsection{Implementation Details}
The details in network architecture and the details in training and testing phase are elaborated as follows.

\subsubsection{\textbf{The Details in Network Architecture}} As for the network, the input images are resized to 224$\times$224. The weights of the the first fourth convolutional blocks in the refined VGG16 are initialized from the origin VGG16, which is trained on the ImageNet. Other weights are randomly sampled by truncated\_normal distribution.
Taking into account the specificity of VAP task, namely including images with high-contrast objects and images with high-semantics objects, both low-level contrast and high-level semantic features should be harnessed. To this end, the outputs of $conv$1-2, $conv$2-2, $conv$3-3, $conv$4-3 and $conv$5-3 are gathered from the refined VGG16. The spatial dimensions of the obtained features are 224$\times$224, 112$\times$112, 56$\times$56, 28$\times$28, and 28$\times$28, respectively. Then, the outputs of $conv$1-2 and $conv$2-2 are fed into the contrast feature extraction block for automatically obtaining contrast features in a deep network. These features have different receptive field and contain various complementary information. Subsequently, these obtained multiple features are input into the next steps for inferring the final visual attention map.

The acquired basic features ${\bf{F}}_{2}$, ${\bf{F}}_{3}$, ${\bf{F}}_{4}$ and ${\bf{F}}_{5}$ are upsampled scale by scale to enlarge the spatial resolution. Then, each upsampled feature is concatenated with higher spatial resolution feature. It is noteworthy that the upsampling operation is in a resize-convolution approach instead of a simply deconvolution approach due to the checkerboard artifacts of uneven overlap during the deconvolution operation. Specifically, the feature map with low spatial resolution is resized to a higher resolution using a nearest-neighbor interpolation method. Subsequently, a 3$\times$3 convolutional layer is followed to further learn the sampling pattern. In addition, when various features are concatenated, the devised reduction-attention block is employed to automatically learn the combination weight due to the different contribution of various features. To verify the effectiveness of this combination approach, an ablation analysis is performed. Firstly, multiple features are concatenated directly after upsampling to 224$\times$224. Secondly, multiple features are concatenated scale by scale in a densely connected manner. Thirdly, the reduction-attention block is added to automatically learn the combination weight for different features. Expectedly, the best prediction result is obtained when the reduction-attention block is added to learn the combination weight. Therefore, the adopted combination strategy is effective.

As for the readout network in inferring five rough visual attention maps, it consists of three stacked reduction-attention blocks to learn interactions across channels instead of learn new spatial features. The output channel numbers of the three stacked reduction-attention block are 32, 16 and 1, respectively. After the readout network, five rough visual attention maps are obtained. Next, to address the centre-bias problem in the human eye fixations, the learned centre-bias prior map is fused with the obtained rough visual attention maps. Then, we perform an image-level fusion by a weighted-fusion layer to get the fused visual attention map, which adopts similar architecture with the readout network. Afterwards, the fused visual attention map is convolved with a 7$\times$7 Gaussian kernel so that we obtain the visual attention map with spatial size of 224$\times$224. Finally, the visual attention map with spatial size of 224$\times$224 is resized to get the same size as the original image.

\begin{figure*}[tp]
\begin{center}
\includegraphics[width=0.8\linewidth]{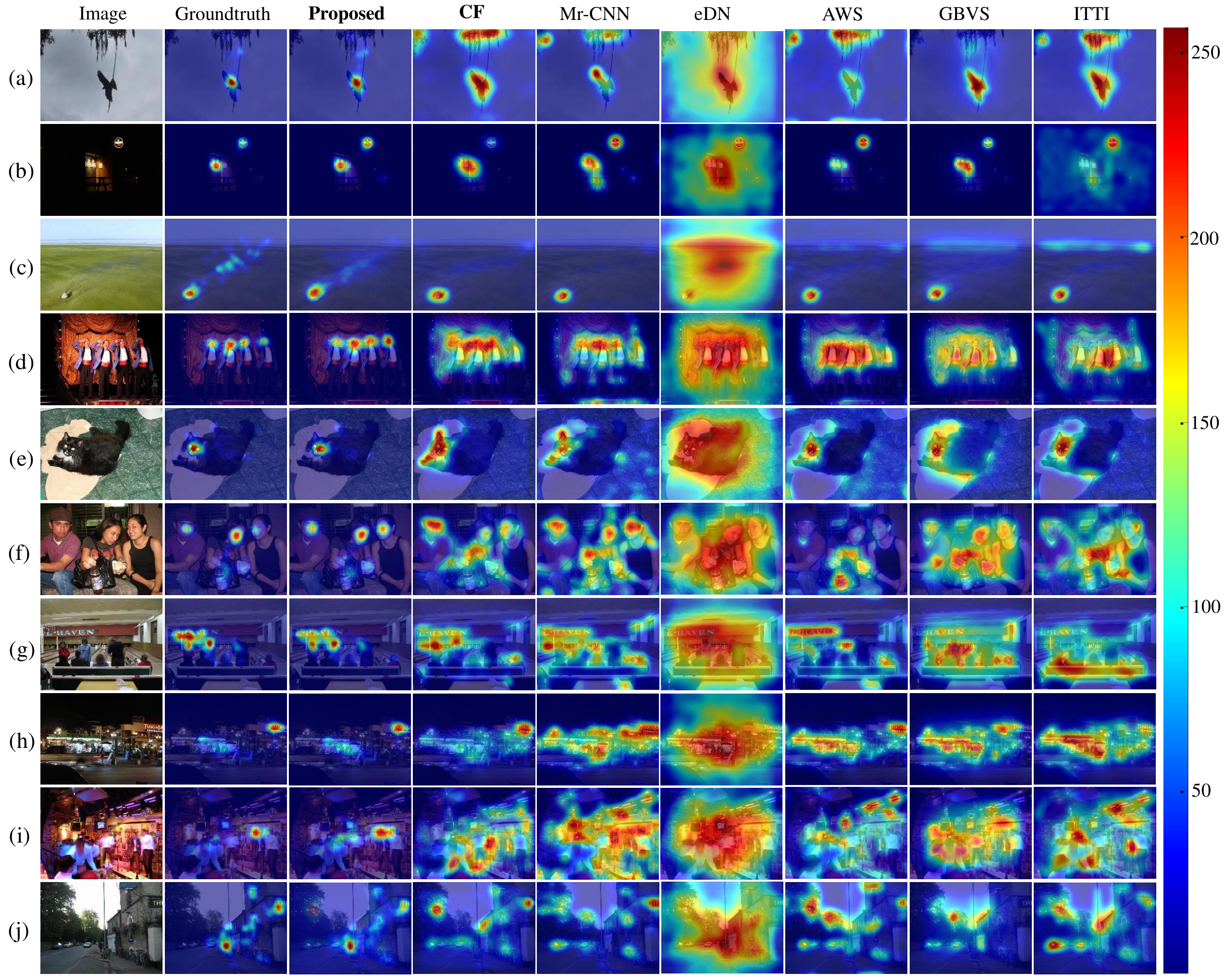}
\renewcommand{\figurename}{Fig.}
\vspace{-5 mm}
\end{center}
    \caption{\small{Qualitative results of different methods on MIT1003 dataset. The first column shows various stimulus images, the second column shows the corresponding groundtruth density map, the third column shows the results of the proposed method ({\bf{Proposed}}), the fourth column shows the results based on the low-level contrast features from the proposed contrast feature extraction block ({\bf{CF}}), and the rest of columns show the results of other five state-of-the-art methods: Mr-CNN \cite{Liu2018Learning}, eDN \cite{Vig2014Large}, AWS \cite{Garcia2012On}, GBVS \cite{Sch2006Graph} and ITTI \cite{Itti2002A}.}}
\label{fig:5}
\end{figure*}

\subsubsection{\textbf{Training and Testing}} The proposed method is trained on SALICON dataset and tested on OSIE , MIT1003, TORONTO and PASCAL-S datasets with the $tensorflow$ library. For the training phase, all input images are resized to a uniform size with 224$\times$224 to satisfy the input size of the VGG network. The minibatch input is fed into the network, and the final loss (Eq. \ref{Eq.:13}) is minimized using a RMSprop optimizer. The initial learning rate is set as $10^{-4}$. The weight decay is set as 0.0005. And the momentum is set as $0.9$. We find that when the batch size is set to 10, the convergence speed is quickest and the performance is best. 5000 validation images of SALICON dataset are exploited to decide when to stop the training process for avoiding overfitting furtherly. Specifically, after the training of each epoch, the performance on these 5000 validation images are evaluated and the training is terminated when the performance begin to decrease or the training epoch achieves 100 to reduce unnecessary training cost. For the testing phase, we use OSIE, MIT1003, TORONTO and PASCAL-S datasets for evaluating the prediction result. Given a query image, it will generate a final visual attention map automatically with the same size as the query image. The experiment is performed on our PC with a TITAN X (Pascal) GPU and 64G RAM.
\subsection{Comparison with State-of-the-arts}
To demonstrate the effectiveness of the proposed method, we compare the proposed method with several state-of-the-art methods. The compared methods are comprehensive, including 6 classical methods, 4 deep learning methods trained on SALICON dataset, 2 deep learning methods trained on MIT1003 dataset and 2 deep learning methods trained on two datasets.

\subsubsection{\textbf{Classical Methods}} These methods are widely used to compare the performance of the VAP model. The comparison classical methods include ITTI \cite{Itti2002A}, JUDD \cite{Judd2010Learning}, BMS \cite{Zhang2014Saliency}, GBVS \cite{Sch2006Graph}, CAS \cite{Goferman2010Context} and AWS \cite{Garcia2012On} methods. ITTI is the seminal work for VAP, in which multi-scale image features are combined into a single topographical for generating visual attention map. JUDD predicts the visual attention map based on low, middle and high-level image features. BMS finds the surrounding area by Boolean topology analysis to predict the visual attention map. GBVS is a bottom-up visual VAP model, which is based on graph. CAS proposes a detection algorithm based on the four principles observed in the psychology literature. AWS utilizes a contextual adaptation mechanism to ensure that the contribution of image points to optical changes is invariant in behavior caused in the visual system.

\subsubsection{\textbf{Methods Trained on SALICON Dataset:}} These methods are similar to the proposed method, which also adopts SALICON dataset for training. The comparison methods trained on SALICON dataset include SAM-VGG \cite{Cornia2016Predicting}, SAM-ResNet \cite{Cornia2016Predicting}, DVA \cite{Wang2018Deep}, and SU \cite{Kruthiventi2016Saliency} methods. SAM-VGG and SAM-ResNet come from one of the latest work for VAP, which highlight the most significant areas of the input image to refine the predictive saliency map by convolution LSTM iteration. DVA is also a recent state-of-the-art work, which can capture hierarchically saliency information from deep layers with global saliency information to shallow layers with local saliency responses. SU capitalizes on a deep convolutional neural network to predict the visual attention map and segment salient objects in a unified framework.

\subsubsection{\textbf{Methods Trained on MIT1003 Dataset}} These methods are used to compare the performance when using different training dataset. The comparison methods trained on MIT1003 dataset include Mr-CNN \cite{Liu2018Learning} and eDN \cite{Vig2014Large} methods. Mr-CNN adopts a multi-resolution convolutional neural network to infer local contrast, global contrast, and top-down visual factors from raw image data simultaneously so as to generate the visual attention map. eDN is the first VAP method based on deep convolutional neural networks.

\subsubsection{\textbf{Methods Trained on Two Datasets:}} These methods are trained in two stages, including a larger dataset-SALICON for first stage training and a smaller dataset for second training. These methods are used to compare the performance when using the augmented training dataset. The comparison methods trained on MIT1003 dataset include JN \cite{Pan2016Shallow} and DeepFix \cite{Sss2017DeepFix} methods. JN proposed two different VAP designs, including a shallow convent trained from scratch, and another deeper solution with the first three layers adapted from another network of trained classifications. DeepFix achieves the best score with respect to most evaluation metrics on several challenging benchmark datasets.

On one hand, the qualitative visual comparisons on MIT1003 dataset are shown in Fig. \ref{fig:5}. Considering that some codes are not public, only parts of visual attention maps are generated using their available codes with recommended parameters settings. Three kind of stimulus images are selected in Fig. \ref{fig:5}, including images with high-contrast object (color, intensity, texture, {\it etc.}): Fig. \ref{fig:5} (a)-(c), images with high-semantic object (face, head, eye, {\it etc.}): Fig. \ref{fig:5} (d)-(f), and images in complex scenes (both objects are included): Fig. \ref{fig:5} (g)-(j). As can be seen, the eDN method achieves the worst result because of its poor semantic features for limited training data and scarce contrast features. For images with high-contrast objects, it can be seen that all methods except for eDN can highlight the regions which attract human attention roughly. This is because these methods utilize low-level features, which contains local contrast information. Note that the results based on the low-level contrast features from the proposed contrast feature extraction block are quite satisfied. For images with high-semantic objects, the traditional methods based on local contrast are not applicable any more. The methods based learning show better results, which is because these methods can learn abundant semantic information layer by layer. For images in complex scenes, we find that the results of the proposed method are excellent. This is because the proposed method can effectively take advantage of low-level contrast features and high-level semantic features based on the unique physiological structure of humans eyes and human prior knowledge. Overall, the proposed method can surpass other listed methods in any case, especially for the images in complex scenes (Fig. \ref{fig:5} (j)-(g)). In addition, some visual attention maps on the other three testing datasets using the proposed method are given in Fig. 6. As it can be seen, the generated visual attention maps using the proposed method are also close to the groundtruth density maps on OSIE, TORONTO and PASCAL-S datasets. Therefore, the results in Fig. 7 can further illustrate the good performance of the proposed method.
\begin{figure}[tp]
\begin{center}
\includegraphics[width=\linewidth]{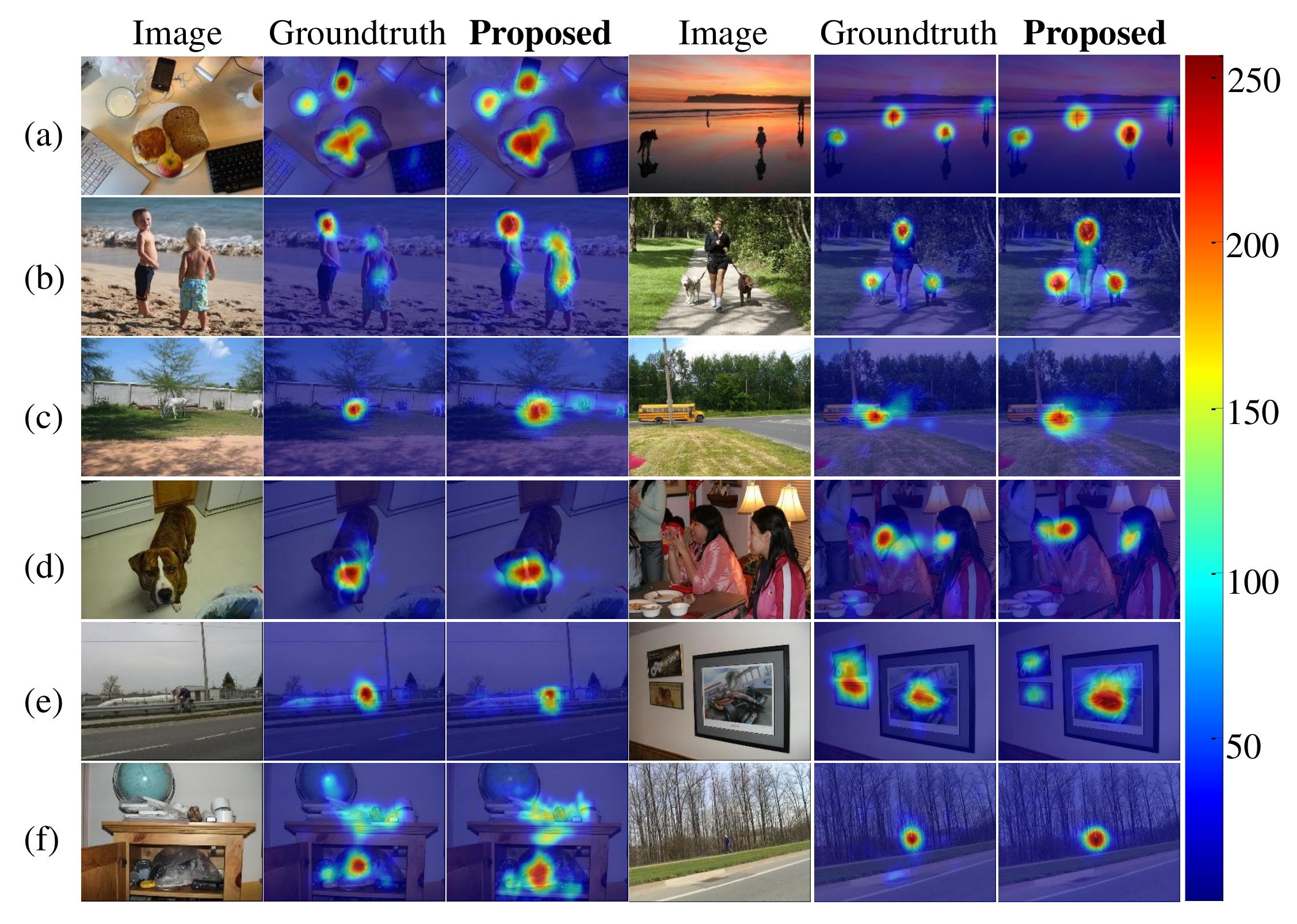}
\renewcommand{\figurename}{Fig.}
\vspace{-8 mm}
\end{center}
    \caption{\small{Some visual attention maps selected from OSIE (a-b), TORONTO (c-d) and PASCAL-S (e-f) datasets. The first column shows various stimulus images, the second column shows the corresponding groundtruth density map, and the third column shows the results of the proposed method ({\bf{Proposed}}).}}
\label{fig:6}
\end{figure}

On the other hand, we also compare our results with several state-of-the-art methods qualitatively on OSIE, TORONTO, PASCAL-S and MIT1003 datasets. Since most VAP methods based on deep learning are not publicly available and the reported results are not based on all considered datasets, the comparison methods are different depending on the dataset. Tables \ref{tab:1}-\ref{tab:4} show the results on OSIE, TORONTO, PASCAL-S and MIT1003 datasets, respectively. As can be seen evidently in these tables, the proposed method surpasses other methods by a substantial margin with respect to most evaluation metrics, especially in s-AUC. The high s-AUC value shows that the proposed method can accurately locate gaze positions of human eyes for the query image. Specifically, on TORONTO dataset (see Table \ref{tab:2}), the proposed method achieves state-of-the-art results. On OSIE and PASCAL-S datasets, the proposed method can also struggle to compete with the most popular DeepFix method. A little imperfection is that the CC of the proposed method is slightly lower than the most popular DeepFix method on OSIE dataset (see Table \ref{tab:1}) and PASCAL-S dataset (see Table \ref{tab:3}). This is because the DeepFix method adopts a more complicated two-stage training manner and more training data, while we only use 10000 training images from SALICON dataset to train the proposed method directly. On MIT1003 dataset (see Table \ref{tab:4}), the proposed method also has a satisfied performance compared to other state-of-the-art methods. In general, the proposed method can achieve promising results, and we mainly attribute the good performance of the proposed method to the simultaneous use of both high-level semantic and low-level contrast features.

\begin{table}
\caption{THE QUANTITATIVE COMPARISON ON OSIE DATASET.}
\begin{center}
\footnotesize
\setlength{\tabcolsep}{1.0mm}{
\begin{tabular}{ccccccc} \toprule
Method    & CC$\uparrow$    & s-AUC$\uparrow$  & AUC-Judd$\uparrow$ & AUC-Borji$\uparrow$ & NSS$\uparrow$ & EMD$\downarrow$ \\ \hline
AWS \cite{Garcia2012On}    & 0.45  & 0.76  & 0.82  & 0.81  & 1.45  & 1.93\\
GBVS \cite{Sch2006Graph}    & 0.44  & 0.68  & 0.82  & 0.80   & 1.35  & 1.67 \\
BMS \cite{Zhang2014Saliency}     & 0.46  & 0.76  & 0.83  & 0.82  & 1.47  & 1.89 \\
\hline
eDN \cite{Vig2014Large}    & 0.40  & 0.68  & 0.82  & 0.82  & 1.16  & 2.02 \\
\hline
DeepFix \cite{Sss2017DeepFix} & \textbf{0.80}  & 0.79  & 0.91  & 0.83  & \textbf{3.04}  & 1.04 \\
\hline
SAM-VGG \cite{Cornia2016Predicting} & 0.78  & 0.70  & 0.91  & 0.80  & 2.74  & 1.12 \\
SAM-Resnet \cite{Cornia2016Predicting} & 0.79  & 0.70  & 0.91  & 0.81  & 2.88  & 0.97 \\
\bf{Proposed}    & 0.77  & \textbf{0.84}  & \textbf{0.92}  & \textbf{0.87}  & 2.87  & \textbf{0.67} \\
\bottomrule
\end{tabular}}
\end{center}
 \label{tab:1}%
\end{table}

\begin{table}
\caption{THE QUANTITATIVE COMPARISON ON TORONTO DATASET.}
\begin{center}
\footnotesize
\setlength{\tabcolsep}{1.5mm}{
\begin{tabular}{cccccc} \toprule
Method      & CC$\uparrow$    & s-AUC$\uparrow$  & AUC-Judd$\uparrow$ & AUC-Borji$\uparrow$ & NSS$\uparrow$ \\ \hline
ITTI \cite{Itti2002A}        & 0.48  & 0.65  & 0.80  & 0.80  & 1.30 \\
GBVS \cite{Sch2006Graph}      & 0.57  & 0.64  & 0.83  & 0.83  & 1.52 \\
JUDD \cite{Judd2010Learning}        & 0.41  & 0.67  & 0.78  & 0.77  & 1.15 \\
CAS \cite{Goferman2010Context}        & 0.45  & 0.69  & 0.78  & 0.78  & 1.27 \\
\hline
eDN \cite{Vig2014Large}        & 0.50  & 0.62  & 0.85  & 0.84  & 1.25 \\
Mr-CNN \cite{Liu2018Learning}     & 0.49  & 0.71  & 0.80  & 0.79  & 1.41 \\
\hline
SAM-VGG \cite{Cornia2016Predicting}        & 0.72  & 0.68  & 0.89  & 0.82  & 1.97 \\
SAM-Resnet \cite{Cornia2016Predicting}        & 0.73  & 0.69  & 0.89  & 0.82  & 2.11 \\
DVA \cite{Wang2018Deep}        & 0.72  & 0.76  & 0.86  & \textbf{0.86}  & 2.12 \\
\bf{Proposed}        & \textbf{0.74}  & \textbf{0.77}  & \textbf{0.90}  & 0.84  & \textbf{2.39} \\
\bottomrule
\end{tabular}}
\end{center}
 \label{tab:2}%
\end{table}

\begin{table}
\caption{THE QUANTITATIVE COMPARISON ON PASCAL-S DATASET.}
\begin{center}
\footnotesize
\setlength{\tabcolsep}{0.6mm}{
\begin{tabular}{ccccccc} \toprule
Method   & CC$\uparrow$    & s-AUC$\uparrow$  & AUC-Judd$\uparrow$ & AUC-Borji$\uparrow$ & NSS$\uparrow$ & EMD$\downarrow$ \\ \hline
GBVS \cite{Sch2006Graph}      & 0.51  & 0.65  & 0.84  & 0.82  & 1.36  & 1.16\\
BMS \cite{Zhang2014Saliency}      & 0.44  & 0.67  & 0.80  & 0.78  & 1.28  & 1.32\\
\hline
eDN \cite{Vig2014Large}      & 0.55  & 0.65  & 0.89  & \textbf{0.87}   & 1.42  & 1.29 \\
\hline
JN \cite{Pan2016Shallow}      & 0.68  & 0.69  & 0.88  & 0.86  & 1.90  & 1.04 \\
DeepFix \cite{Sss2017DeepFix}  & \textbf{0.78}  & 0.73  & 0.91  & 0.82  & 2.60  & 0.54 \\
\hline
SU \cite{Kruthiventi2016Saliency}       & 0.69  & 0.72  & 0.89  & 0.81  & 2.22  & 0.73 \\
SAM-VGG \cite{Cornia2016Predicting}       & 0.74  & 0.68  & 0.90  & 0.80  & 2.56  & 0.98 \\
SAM-Resnet \cite{Cornia2016Predicting}       & 0.0.75  & 0.68  & 0.91  & 0.80  & 2.70  & 0.86 \\
DVA \cite{Wang2018Deep}      & 0.66  & 0.77  & 0.89  & 0.85  & 2.26  & $-$  \\
\bf{Proposed}      & 0.74  & \textbf{0.81}  & \textbf{0.93}  & 0.85  & \textbf{2.87}  & \textbf{0.53} \\
\bottomrule
\end{tabular}}
\end{center}
 \label{tab:3}%
\end{table}

\begin{table}
\caption{THE QUANTITATIVE COMPARISON ON MIT1003 DATASET.}
\begin{center}
\footnotesize
\setlength{\tabcolsep}{1.0mm}{
\begin{tabular}{ccccccc} \toprule
Method   & CC$\uparrow$    & s-AUC$\uparrow$  & AUC-Judd$\uparrow$ & AUC-Borji$\uparrow$ & NSS$\uparrow$  \\ \hline
ITTI \cite{Itti2002A}         & 0.33  & 0.66  & 0.77  & 0.76  & 1.10 \\
GBVS \cite{Sch2006Graph}      & 0.42  & 0.66  & 0.83  & 0.81  & 1.38\\
JUDD \cite{Judd2010Learning}      & 0.30  & 0.68  & 0.76  & 0.74  & 1.02 \\
BMS \cite{Zhang2014Saliency}      & 0.36  & 0.69  & 0.79  & 0.76  & 1.25  \\
\hline
eDN \cite{Vig2014Large}      & 0.41  & 0.66  & 0.85  & 0.84   & 1.29  \\
Mr-CNN \cite{Liu2018Learning}      & 0.38  & 0.73  & 0.80  & 0.77  & 1.36  \\
\hline
DeepFix \cite{Sss2017DeepFix}  & 0.72  & 0.74  & 0.90  & 0.87  & 2.58  \\
\hline
SAM-VGG \cite{Cornia2016Predicting}       & 0.76  & 0.61  & 0.91  & 0.85  & 2.85   \\
SAM-Resnet \cite{Cornia2016Predicting}       & \textbf{0.77}  & 0.62  & 0.91  & 0.86  & \textbf{2.89}   \\
DVA \cite{Wang2018Deep}      & 0.64  & 0.77  & 0.87  & 0.85  & 2.38   \\
\bf{Proposed}      & 0.73  & \textbf{0.79}  & \textbf{0.92}  & \textbf{0.88}  & 2.54   \\
\bottomrule
\end{tabular}}
\end{center}
 \label{tab:4}%
\end{table}

\subsection{Ablation Analysis}
In this subsection, we analyze the effectiveness of each component for the proposed method. We perform the experiment with different design options and give an ablation analysis for each component. More specifically, as is shown in Table \ref{tab:5}, seven different variations are constructed: the fusion feature of F1 and F2 without the contrast feature extraction block (NCF) method, the Contrast Feature (CF) method via combining the contrast features F1 and F2 with the Contrast Feature Extraction Block directly, the Semantic Feature (SF) method by merging the semantic features F3, F4 and F5 directly, the Directly Combining Feature (DCF) method by concatenating all of contrast features and semantic features directly, the Densely Combining Feature (DenCF) method using the designed densely connected connected block to combine all features, the Densely Combining Feature with the learnable Center-Bias Prior (DenCF+CBP) method, and the final version of the proposed method by adding the designed reduction-attention block to the DenCF method. Each component of the proposed method is analyzed in the following aspects.
\begin{table}
  \centering
\scriptsize
  \caption{THE ABLATION STUDY ON OSIE, TORONTO, PASCAL-S AND MIT1003 DATASETS.}
  \begin{threeparttable}
 \setlength{\tabcolsep}{1.0mm}{	
    \begin{tabular}{ccccccccc}\toprule
     Dataset &Method & s-AUC$\uparrow$ & $\Delta$s-AUC &NSS$\uparrow$ & $\Delta$NSS & CC$\uparrow$ & $\Delta$CC \\
    \hline
    \multirow{7}*{OSIE}   &NCF             & 0.701     & -0.143     & 1.212     & -1.661     & 0.492     & -0.280 \\
                         ~&CF              & 0.734     & -0.110     & 1.953     & -0.920     & 0.594     & -0.178 \\
                         ~&SF              & 0.796     & -0.048     & 2.494     & -0.379     & 0.712     & -0.060 \\
                         ~&DCF             & 0.812     & -0.032     & 2.696     & -0.177     & 0.734    & -0.029 \\
                         ~&DenCF           & 0.834     & -0.010     & 2.810     & -0.063     & 0.746     & -0.026 \\
                         ~&DenCF+CBP       & 0.820     & -0.024     & 2.851     & -0.022     & 0.761     & -0.011 \\
                         ~&\bf{Proposed}            & \textbf{0.844}    & $-$      & \textbf{2.873}  & $-$    & \textbf{0.772}     & $-$ \\
    \hline
    \multirow{7}*{TORONTO}&NCF             & 0.623     & -0.143     & 1.376     & -1.016     & 0.441     & -0.302 \\
                         ~&CF              & 0.652     & -0.114     & 1.905     & -0.487     & 0.566     & -0.177 \\
                         ~&SF              & 0.716     & -0.050     & 2.249     & -0.143     & 0.681     & -0.062 \\
                         ~&DCF             & 0.732     & -0.034     & 2.301     & -0.091     & 0.717    & -0.026 \\
                         ~&DenCF           & 0.746     & -0.020     & 2.371     & -0.021     & 0.729     & -0.014 \\
                         ~&DenCF+CBP       & 0.739     & -0.027     & 2.356     & -0.036     & 0.720     & -0.023 \\
                         ~&\bf{Proposed}            & \textbf{0.766}    & $-$      & \textbf{2.392}  & $-$    & \textbf{0.743}     & $-$ \\
                         \hline
    \multirow{7}*{PASCAL-S}&NCF            & 0.692     & -0.120     & 1.386     & -1.481     & 0.432     & -0.310 \\
                         ~&CF              & 0.710     & -0.102     & 2.016     & -0.851     & 0.602     & -0.140 \\
                         ~&SF              & 0.784     & -0.028     & 2.595     & -0.272     & 0.689     & -0.053 \\
                         ~&DCF             & 0.791     & -0.021     & 2.712     & -0.155     & 0.706     & -0.036 \\
                         ~&DenCF           & 0.801     & -0.011     & 2.785     & -0.082     & 0.715     & -0.027 \\
                         ~&DenCF+CBP       & 0.795     & -0.017     & 2.813     & -0.054     & 0.719     & -0.023 \\
                         ~&\bf{Proposed}            & \textbf{0.812}    & $-$      & \textbf{2.867}  & $-$    & \textbf{0.742}     & $-$ \\
    \hline
    \multirow{7}*{MIT1003}&NCF             & 0.685     & -0.109     & 1.056     & -1.485     & 0.452     & -0.281 \\
                         ~&CF              & 0.702     & -0.092     & 1.752     & -0.789     & 0.594     & -0.139 \\
                         ~&SF              & 0.717     & -0.077     & 2.251     & -0.290     & 0.659     & -0.074 \\
                         ~&DCF             & 0.735     & -0.059     & 2.453     & -0.088     & 0.681     & -0.052\\
                         ~&DenCF           & 0.758     & -0.036     & 2.510     & -0.031     & 0.706     & -0.027 \\
                         ~&DenCF+CBP       & 0.756     & -0.038     & 2.532     & -0.009     & 0.712     & -0.021 \\
                         ~&\bf{Proposed}            & \textbf{0.7694}    & $-$      & \textbf{2.5412}  & $-$    & \textbf{0.733}     & $-$ \\
\bottomrule
    \end{tabular}}%
  \label{tab:5}%
\end{threeparttable}

\end{table}%

\subsubsection{\textbf{Low-level Contrast Features}} To validate the contribution of the extracted low-level contrast features, we compare the three methods CF, SF and DCF. As is shown in Table \ref{tab:5}, the DCF method outperformed the SF method in all listed metrics. This is because the low-level contrast features are considered in the DCF method compared to the SF method. Therefore, we can conclude that the low-level contrast features do contribute to VAP, and the low-level contrast features are complementary to the high-level semantic features by a substantial margin.

\subsubsection{\textbf{Combination Strategy}} To consider the advantages of multiple features, these features should be exploited simultaneously. Here, a densely connected manner is adopted to well combine these features. This combination strategy can further excavate the feature representation of each level and utilize the advantage of multiple features simultaneously. To further demonstrate the effectiveness of our combination strategy, we compared the performance of the proposed method with directly combined feature method. In Table \ref{tab:5}, we can clearly see that the DenCF method is better than the DCF method, which manifests the effectiveness of the densely strategy.

\subsubsection{\textbf{Designed Centre-bias Prior}} A learnable centre-bias block is designed for generating a data-dependent centre-bias prior map, which can compensate for the centre-bias when human observing images and improve the predicted results. We experiment with the designed centre-bias prior block (DenCF+CBP method in Table \ref{tab:5}) and without the designed centre-bias prior block (DenCF method in Table \ref{tab:5}) respectively for validating the effectiveness of the designed centre-bias prior block. The results show that the DenFF+CBP method precedes the DenCF method, which indicates the designed centre-bias prior block can effectively alleviate the centre-bias problem and improve the predicted results.

\subsubsection{\textbf{Reduction-Attention Block}} The reduction-attention block is devised to adaptively recalibrate channel-wise feature responses during bio-inspired representation learning. We performed the experiment with this block (Proposed in Table \ref{tab:5}) and without this block (DenCF+CBP method in Table \ref{tab:5}) to demonstrate its effectiveness. Obviously, the proposed method adds an improvement of 0.027, 0.036 and 0.023 On TORONTO dataset in terms of s-AUC, NSS and CC, respectively, which proves the effectiveness of the reduction-attention block.

\subsubsection{\textbf{Contrast Feature Extraction Block}} The contrast feature extraction block is proposed for low-level contrast features automatical extraction in a deep network. We performed the experiment with this block (CF in Table \ref{tab:5}) and without this block (NCF method in Table \ref{tab:5}) to demonstrate its effectiveness. As it can be seen, the performance obtained by CF method is better than NCF method on the CC, AUC and NSS metrics, which manifests the proposed contrast feature extraction block has a profound effect on the generation of the final visual attention map.

\subsection{Timing}
According to \cite{He2015Convolutional}, the time complexity of each convolutional and pooling layer can be computed as $O(M^{2}\cdot K^{2}\cdot C_{in}\cdot C_{out})$ , where $M$ denotes the size of the output feature map, $K$ represents the size of the convolutional kernel, $C_{in}$ is the channel number of the input feature map, and $C_{out}$ indicates the channel number of the output feature map. As for the fully connected layer, it can be considered as a special convolutional layer, in which the size of the output feature map and the convolutional kernel are 1. As a result, the time complexity of each fully connected layer can be computed as $o(C_{in}\cdot C_{out})$. With regard to the element-wise multiplication in reduction-attention block, its time complexity is $O(M^{2})$. Considering that other operation (bias operation, concatenation operation, etc.) can be ignored, the time complexity of proposed network is $O(M^{2}\cdot K^{2}\cdot C_{in}\cdot C_{out})+O(C_{in}\cdot C_{out})+O(M^{2})$.

In addition, according to \cite{Wang2018Deep}, a summary of these methods is provided in Table \ref{tab:6}. As visible, most of existing VAP methods are off-line training or based on deep learning framework. Since some codes of deep learning methods are not public, we only report the inferring speed performance of DVA, SalNet, Mr-DNN, and eDN with other non-deep learning methods. Obviously, the inferring speed of the proposed method can struggle to compete with DVA and SalNet, although the proposed method adopts a slightly smaller input size than DVA and SalNet.
\begin{table}
  \centering
\footnotesize
  \caption{A SUMMARY OF THE COMPUTATION TIME FOR THE PROPOSED METHOD AND 9 STATE-OF-THE-ART VAP METHODS.}
  \begin{threeparttable}
 \setlength{\tabcolsep}{1.3mm}{
    \begin{tabular}{ccccc}\toprule
     Method         & Input Size                & Training      &Deep Learning          &Runtime \\
     \bottomrule
    \specialrule{0em}{1pt}{1.2pt}
    ITTI \cite{Itti2002A}           & full size                 & No            & No                    & 4s\\
    \specialrule{0em}{1pt}{1.3pt}
    GBVS \cite{Sch2006Graph}           & full size                 & No            & No                    & 2s\\
    \specialrule{0em}{1pt}{1.2pt}
    CAS \cite{Goferman2010Context}            & max\{w,h\}=250            & No            & No                    & 16s\\
    \specialrule{0em}{1pt}{1.2pt}
    BMS \cite{Zhang2014Saliency}            & w=600                     & No            & No                    & 0.3s\\
    \specialrule{0em}{1pt}{1.2pt}
    JUDD \cite{Judd2010Learning}           & 200$\times$200            & Yes           & No                    & 10s\\
    \specialrule{0em}{1pt}{1.2pt}
    Mr-CNN \cite{Liu2018Learning}         & 400$\times$400            & Yes           & Yes                   & 14s$\star$ \\
    \specialrule{0em}{1pt}{1.2pt}
    SalNet \cite{Pan2016Shallow}          & 320$\times$240            & Yes           & Yes                   & 0.1s$\star$ \\
     \specialrule{0em}{1pt}{1.2pt}
    DVA \cite{Wang2018Deep}             & max\{w,h\}=256            & Yes           & Yes                   & 0.1s$\star$ \\
    \specialrule{0em}{1pt}{1.5pt}
    \bf{Propsed}         & 224$\times$224            & Yes           & Yes                   & 0.1s$\star$ \\
\bottomrule
    \end{tabular}}%
    \begin{tablenotes}
\item$\star$  Runtime with GPU.
\end{tablenotes}
  \label{tab:6}%
\end{threeparttable}

\end{table}%

\begin{figure}[htb]
\begin{center}
\includegraphics[width=\linewidth]{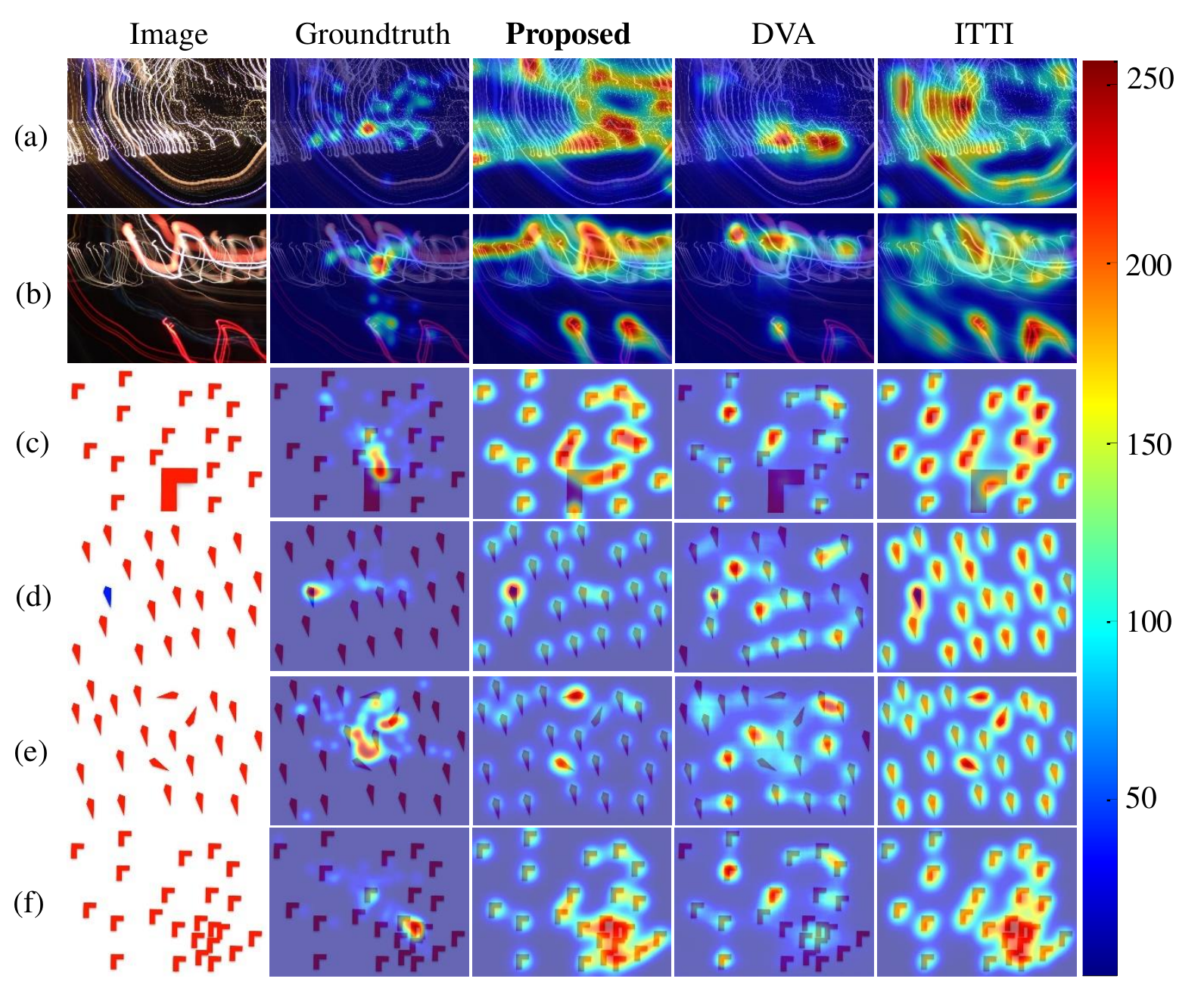}
\renewcommand{\figurename}{Fig.}
\vspace{-8 mm}
\end{center}
    \caption{\small{Failure cases selected from MIT1003 dataset. (a)-(b) are  the images without the high-contrast and high-semantics information. (c)-(f) are the synthetic images.}}
\label{fig:7}
\end{figure}
\subsection{Failure Cases Analysis}
Some failure predictions of the proposed method have been shown in Fig. \ref{fig:7}. In general, these failure cases can be divided into two categories: 1) the images without the high-contrast and high-semantics information (Fig. \ref{fig:7} (a)-(b)); and 2) the synthetic images (Fig. \ref{fig:7} (c)-(f)). As for the images without the high-contrast and high-semantics information, our method can not perform well maybe because these images have no obvious saliency aspects to be predicted. In addition, the proposed method is difficult to process synthetic images, which may be because the amount of synthetic image is small in the training dataset and the synthetic image contains little semantic information. Although the proposed method performs poorly for the synthetic image, it can still outperform the DVA method, which is entirely based on the deep semantic features. What's more, the performance of the proposed model is similar to the ITTI method, which may be because both the proposed method and the ITTI method consider the low-level contrast features.

\section{Conclusions}  \label{conclutions}

In this paper, an intuitive method has been proposed to generate visual attention map via bio-inspired representation learning. To perform the {\it Visual Attention Prediction} (VAP), the proposed method extracts the low-level contrast features automatically in a deep network, then combines them with the high-level semantic features via bio-inspired representation learning, and finally generates the ultimate visual attention map based on the obtained features after bio-inspired representation learning. Experiment results on MIT1003, OSIE, TORONTO and PASCAL-S datasets demonstrate that the designed contrast feature extraction block which is proposed to learn low-level contrast features in a deep network is effective. Moreover, other components in the proposed method are proved to be effective. In addition, the proposed method achieves the superior performance compared with other state-of-the-art methods.

\bibliographystyle{IEEEtran}
\bibliography{reference}
\end{document}